\pdfoutput=1

\documentclass[11pt]{article}

\usepackage[preprint]{acl}

\usepackage{times}
\usepackage{latexsym}

\usepackage[T1]{fontenc}

\usepackage[utf8]{inputenc}

\usepackage{microtype}

\usepackage{inconsolata}

\usepackage{graphicx}
\usepackage{booktabs}
\usepackage{tabularx}
\usepackage{geometry}
\usepackage{amssymb} 
\usepackage{pifont} 
\usepackage{times}
\usepackage{latexsym}
\usepackage{subcaption}
\usepackage{rotating}
\usepackage{mdframed}
\usepackage{array}
\usepackage{authblk}

\title{Are Small Language Models Ready to Compete with Large Language Models for Practical Applications?}

\author[1]{\textbf{Neelabh Sinha}}
\author[2\thanks{Work does not relate to position at Meta.}]{\textbf{Vinija Jain}}
\author[3\thanks{Work does not relate to position at Amazon.}]{\textbf{Aman Chadha}}

\affil[1]{Georgia Institute of Technology}
\affil[2]{Meta AI}
\affil[3]{Amazon GenAI}
\affil[ ]{\texttt{nsinha68@gatech.edu}, \texttt{hi@vinija.ai}, \texttt{hi@aman.ai}}

\begin{document}
\maketitle

\begin{abstract}
The rapid rise of Language Models (LMs) has expanded their use in several applications. Yet, due to constraints of model size, associated cost, or proprietary restrictions, utilizing state-of-the-art (SOTA) LLMs is not always feasible. With open, smaller LMs emerging, more applications can leverage their capabilities, but selecting the right LM can be challenging as smaller LMs don't perform well universally. This work tries to bridge this gap by proposing a framework to experimentally evaluate small, open LMs in practical settings through measuring semantic correctness of outputs across three practical aspects: \textit{task types}, \textit{application domains} and \textit{reasoning types}, using diverse prompt styles. It also conducts an in-depth comparison of 10 small, open LMs to identify best LM and prompt style depending on specific application requirement using the proposed framework. We also show that if selected appropriately, they can outperform SOTA LLMs like DeepSeek-v2, GPT-4o-mini, Gemini-1.5-Pro, and even compete with GPT-4o.~\footnote{{GitHub repository containing the code implementation of this work: \href{https://github.com/neelabhsinha/lm-application-eval-kit}{https://github.com/neelabhsinha/lm-application-eval-kit}}}
\end{abstract}

\section{Introduction}

The field of NLP has advanced significantly with the rapid development of Language Models (LMs)~\cite{gpt3,llama2,falcon,gemma,deepseekv2}, which has expanded their use across numerous types like Title Generation~\citep{keles-bayrakli-2024-llama-2}, Data Exploration~\citep{ma-etal-2023-insightpilot}, Dialogue act recognition~\citep{qiang-etal-2024-prompt}; domains like Economics \& Finance~\citep{rajpoot-etal-2024-adapting-llm, yu-etal-2023-harnessing}, Politics~\citep{feng-etal-2023-pretraining}, Nutrition \& Food~\citep{yang2024chatdiet}, News~\citep{kuila-sarkar-2024-deciphering}; and reasoning types~\citep{huang-chang-2023-towards} like ANALOGICAL~\citep{wijesiriwardene-etal-2023-analogical} and Multi-hop~\citep{pan-etal-2021-unsupervised} reasoning.

\begin{figure}
    \centering
    \includegraphics[width=\linewidth]{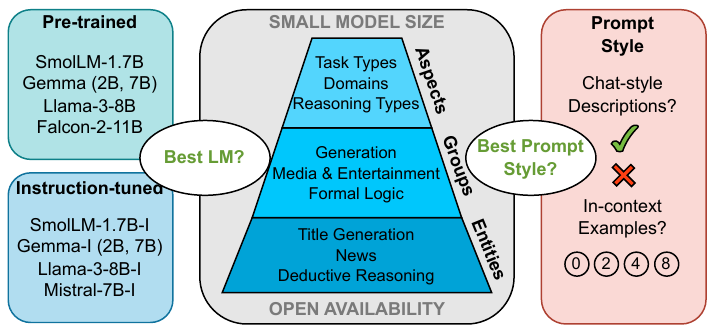}
    \caption{Outline of this work: Performance analysis of open, small-scale LMs and best prompt style for task types, application domains, and reasoning types.}
    \label{fig:outline}
\end{figure}

Despite the growing variety of LMs, their usage in downstream applications is heavily skewed towards limited ones. Analyzing around 50 papers from 2024, we found that while 82.3\% of methods utilized GPT-family LMs, only 41.1\% used Llama variants, and less than 11.8\% experimented with other alternatives like Mistral~\citep{mistral} and Falcon~\citep{falcon}. Some studies also report issues like garbage output and hallucinations~\citep{alhamed-etal-2024-using}, but domain experts often lack the tools to address them effectively through informed LM choice, or correct ways to prompt them. Even the methods that experiment with multiple LMs often select models without a strong motivation~\citep{kuila-sarkar-2024-deciphering}. 

Apart from performance, many of the new LMs are smaller in size, and openly available. Despite the undeniable success of large, proprietary LMs like GPT-4~\citep{gpt4} and Llama-2 70B~\citep{llama2}, their inaccessibility due to limited API access, high costs~\citep{jimenez-gutierrez-etal-2022-thinking}, concerns around data privacy (for GPT), and massive computational demands~\citep{ding2024efficiency} (for Llama) pose significant barriers of usage. Small, open LMs can navigate around those, and also provide additional benefits like on-device usage, faster inference time, data privacy, easier compliance and security management, and low-cost maintainance. For many practitioners -— especially those in research, startups, or sectors with limited resources or high security risk -- leveraging these presents an appealing alternative for functional, financial, or business reasons.

But these new, small, LMs vary a lot in terms of training data, pre-training strategies, and architectural decisions. Additionally, they may not perform globally well like SOTA LLMs due to limitations of scale~\citep{scalinglaws}. Utilization strategies of LMs in inference pipelines can also differ, like zero-shot usage, customizing pre-trained models (e.g., fine-tuning~\citep{mosbach-etal-2023-shot}), using in-context learning~\citep{icl,icl_survey}, prompt engineering~\citep{gpt3}. Writing effective prompts also requires time and domain expertise. So, users need to conduct thorough analysis before choosing the right LM and usage strategy within constraints of time, money, computational resources, which is a complicated task. Although technical reports of some LMs~\citep{gemma,gemma2} provide some insights, not all of them capture real-world, practical scenarios. Therefore, there is a need for a comprehensive practical evaluation framework which can enable determining capabilities of LMs in multiple practical applications, and effective ways to prompt them.

To bridge this gap, we propose a comprehensive framework for evaluating LMs in practical settings along three aspects: task types, application domains, and reasoning types. For each aspect, we select 12, 12, and 10 entities in English, grouping similar ones (e.g., 'Social Media' and 'News' under 'Media and Entertainment'). This three-tier structure (aspect, group, entity) helps identifying patterns in LM capabilities across multiple levels. Using Super-Natural Instructions~\citep{super_natural_instructions}, a meta-dataset encompassing various NLP benchmarks, we evaluate LMs on task instances within this framework. LM usage strategies vary significantly -- ranging from fine-tuning~\citep{mosbach-etal-2023-shot}, PEFT~\citep{peft_survey} or direct usage with/without prompt engineering. Thus, we assess semantic correctness of outputs as an indicator of LMs' inherent abilities, evaluating five pre-trained and five instruction-tuned (IT)~\citep{instructGPT} models across eight prompt styles. Our results show that with careful selection, impact of scale can be reduced. Correctly chosen small, open LM can rival and even outperform models like GPT-4o-mini, GPT-4o~\citep{gpt4o}, DeepSeek-v2~\citep{deepseekv2}, and Gemini-1.5-Pro~\citep{gemini-1.5}, while providing additional benefits. We also evaluate LMs with paraphrases of task definitions to show that results are robust against dataset-induced biases.

In this work, we aim to address these research questions: \textbf{(i)} Can small, open LMs compete with large, proprietary LMs in practical usage? \textbf{(ii)} What can be an exhaustive evaluation framework to conduct this analysis? \textbf{(iii)} For different application needs, how do current best small, open LMs perform in comparison, and which LM is the best choice? \textbf{(iv)} What type of prompt style should be used to extract best results from these LMs? 

Consistent with Figure~\ref{fig:outline}, we make the following \textbf{key contributions}: 

{(i)} Propose a three-tier evaluation framework to analyze performance of LMs for different \textit{task types}, \textit{application domains} and \textit{reasoning types}.

{(ii)} Conduct an in-depth experimental analysis of semantic correctness of outputs of 10 open, small LMs in 1.7B--11B size based on the framework.

{(iii)} Show that appropriate selection of open, small LMs can lead to outperforming SOTA LLMs like GPT-4o-mini, Gemini-1.5-Pro, and competing with GPT-4o.

{(iv)} Compare the performance of LMs with eight prompt styles and recommend the best alternative.

\section{Evaluation Framework}
\label{sec:exp_setup}

We begin with describing our evaluation framework discussing dataset, prompt styles, selection process of aspects, evaluation metrics and experiments.

\subsection{Experimental Dataset}
\label{sec:dataset}

\begin{figure}[htbp]
    \centering
    \includegraphics[width=\linewidth]{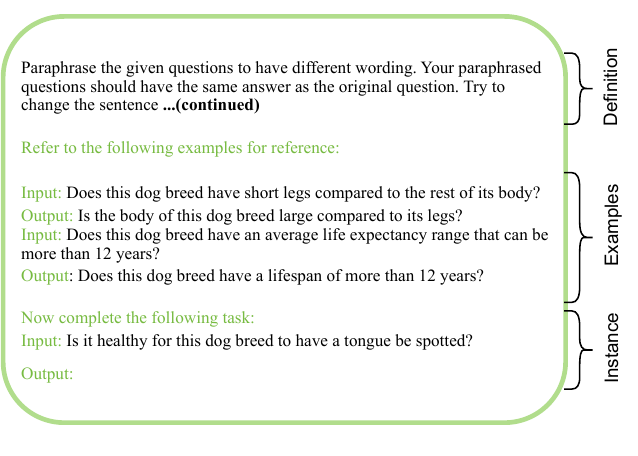}
    \caption{Example of a prompt with definition and 2 examples (text in Green is static text, and others are taken from the dataset).}
    \label{fig:prompt_example}
\end{figure}

We derive our experimental dataset from Super-Natural Instructions~\citep{super_natural_instructions}, which is not a single dataset but a meta-dataset constructed by combining many standard NLP datasets. In addition to the source datasets, it also has definition describing a task in chat-style instruction form and many in-context examples (refer Figure~\ref{fig:prompt_example} for an example) curated by experts. Using datasets from here benefits us by allowing evaluation with various prompt styles and using chat-style instructions -- the way users practically interact with LMs. It also provides labels of task type describing nature of a task (eg. question answering, data to text), domain describing the field of the task (eg. history, news), and reasoning type, describing the type of reasoning (if any) needed in the task (eg. multihop, analogical, etc.), which we also use.

We pick the test split of the dataset for which input and output is English, since most LMs are optimized for that, giving 119 tasks. To avoid redundancy but still take sufficient samples, we take 100 instances per tasks at maximum. Finally, we get 11810 task instances belonging to 12 task types, 36 domains and 18 reasoning types.

\subsection{Prompt Styles}
\label{sec:prompts}

We conduct our experiments using multiple prompt styles - including/excluding chat-style task definitions, and with 0, 2, 4, 8  in-context examples for each instance. Examples help LMs (even pre-trained) with in-context learning~\citep{icl,icl_survey} without altering their parameters. This is followed by an actual task instance. We select examples from positive examples section of the task. This gives 8 prompt styles per task instance. An example of prompt with definition and 2 examples is given in Figure~\ref{fig:prompt_example}. `Input' and `Output' is used since they are universal for all tasks.

\subsection{Selection of Aspects}
\label{sec:selection_of_aspects}

\begin{figure*}[!htb]
    \centering
    \includegraphics[width=\textwidth]{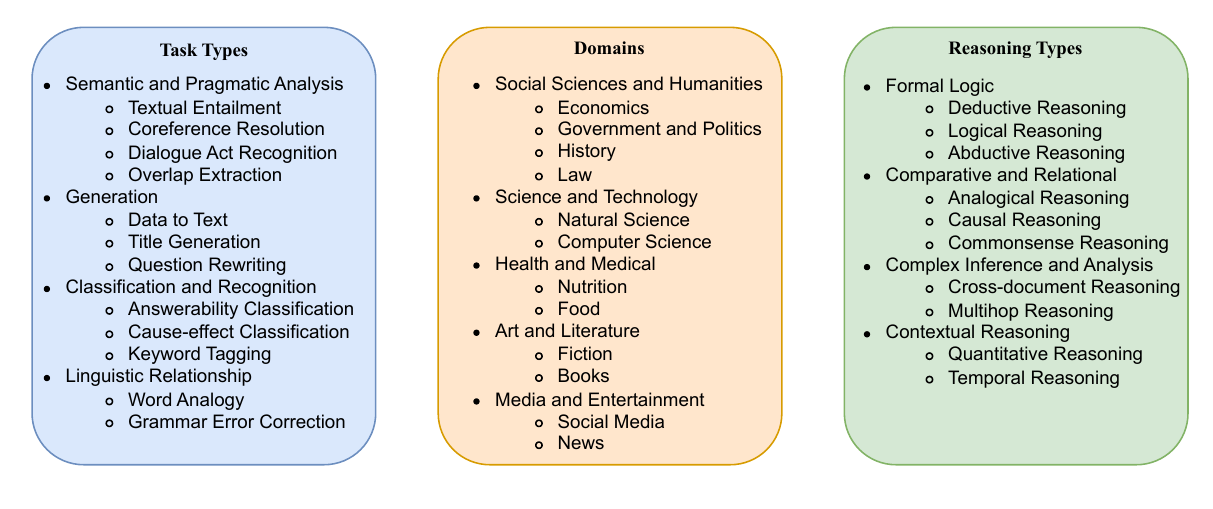}
    \caption{Sections and entities for which the performance of Language Models (LMs) is analyzed (each aspect is divided into groups (like formal logic), and the groups are divided into individual entities (like deductive reasoning). This three-level categorization allows analysis of performance across multiple hierarchies.}
    \label{fig:categories}
\end{figure*}

From the dataset, we divide each task instance into three aspects -- task types, application domains and reasoning types. Since there were many instances for each entity, we filter and rearrange these to create a filtered set for brevity. Our objective was to cover a wide range of application area in each aspect. Therefore, first, we took all the 12 task types in the test set. Among them, for 36 domains and 18 reasoning types, we discarded subsets, very closely similar entities, or ones which didn't have many examples. For example, there were two domains `Computer Science' and `Coding', so we included only Computer Science as Coding can be considered a subset; among the two types of reasoning called `Numerical' and `Quantitative', we included only Quantitative since they were very similar, and so on. As the number of entities were not too many, we did this manually. We always included the more wider scoped entity when resolving these clashes. After taking a broad enough spectrum in all 3 aspects, we constructed groups in each entity and placed them to create a second-level hierarchy, with similar entities in same groups. Our final structure is shown in Figure~\ref{fig:categories}. Here, Domains is an aspect, Social Sciences and Humanities is a group which contains 4 entities, Economics being one of them. Our intention with this is to provide a structure to this study and cover a broad spectrum of entities. Some of the definitions, specifically for reasoning types, are detailed more in a survey~\citep{evaluation_llm_survey} and the dataset repository\footnote{\href{https://instructions.apps.allenai.org}{https://instructions.apps.allenai.org}}.

This allows analysis at three levels of hierarchy - aspect, group and entity level, which is how we address them in rest of this paper. Some tasks can overlap between entities of same aspect~\citep{kuila-sarkar-2024-deciphering} or different aspects~\citep{keles-bayrakli-2024-llama-2}, and some may not belong to any aspect. There are more entities not included here for brevity but listed and evaluated in Appendix~\ref{app:aspects_lm_performance} with dataset statistics.

\subsection{Evaluation Metrics}
\label{sec:evaluation_metrics}

As per the analysis of recent works~\citep{sai-etal-2021-perturbation, xiao-etal-2023-evaluating-evaluation}, evaluating LM outputs using n-gram metrics like ROUGE~\citep{lin2004rouge}, METEOR~\citep{banerjee2005meteor}, etc., have limitations in terms of coherence, consistency, relevance, and fluency. These works also show that BERTScore-recall~\citep{zhang2019bertscore} limits this to a great extent. To be consistent, we evaluate LM's knowledge via semantic correctness of outputs using BERTScore~\citep{zhang2019bertscore} recall with \texttt{roberta-large}~\citep{roberta}.

Some tasks, like classification, aren't generation tasks, but we still consider them as one since they give a uniform evaluation paradigm. By aligning outputs using fine-tuning/ICL~\citep{zhao-etal-2023-simple}, verbalizers~\citep{hu-etal-2022-knowledgeable}, post-processing, labels can be obtained from language outputs.

\subsection{Language Models Used}
\label{sec:models}

The focus for this work is on open LMs from 1.7--11B parameters for adaptability and computational efficiency. Analysis of pre-trained models, trained for next-word prediction, will give an insight into LMs' ability and knowledge to perform the tasks. They can either be used directly or adapted/aligned further. IT models will suit out-of-the-box usage on chat-style human-like instructions due to a simple use-case or unavailability of sufficient data/resources to customize the models.

To cover a broad range of SOTA small, open LMs across sizes, families, our experiments utilize Gemma-2B, Gemma-7B~\citep{gemma}, Llama-3-8B~\cite{llama2,llama3}, Mistral-7B-v0.3~\cite{mistral}, and Falcon-2-11B~\cite{falcon,falcon2}. We also take their instruction-tuned (IT) versions (except Falcon-2-11B - not available). But, we omit Mistral-7B pre-trained from discussion as its results weren't competitive, and Gemma-2 series~\citep{gemma2} since their performance was below Gemma. Model and implementation details are discussed more in Appendix~\ref{app:lm_level_results}, ~\ref{app:implementation_details}. In this paper, suffix "-I" indicates instruction-tuned.
\section{Experiments and Results}
\label{sec:exp_results}

We use all the prompt styles with each of the task instance, do a forward pass on the LM, and decode the output using greedy decoding, which is evaluated with available references. We used greedy as it's reproducible, also other sampling techniques~\citep{top_p_sampling} didn't give any improvement (refer Appendix~\ref{app:comparison_decoding}). The following subsections discusses findings.

\subsection{Performance Correlation of LMs}
\label{sec:overall_performance}

\begin{figure}
    \centering
    \includegraphics[width=\linewidth]{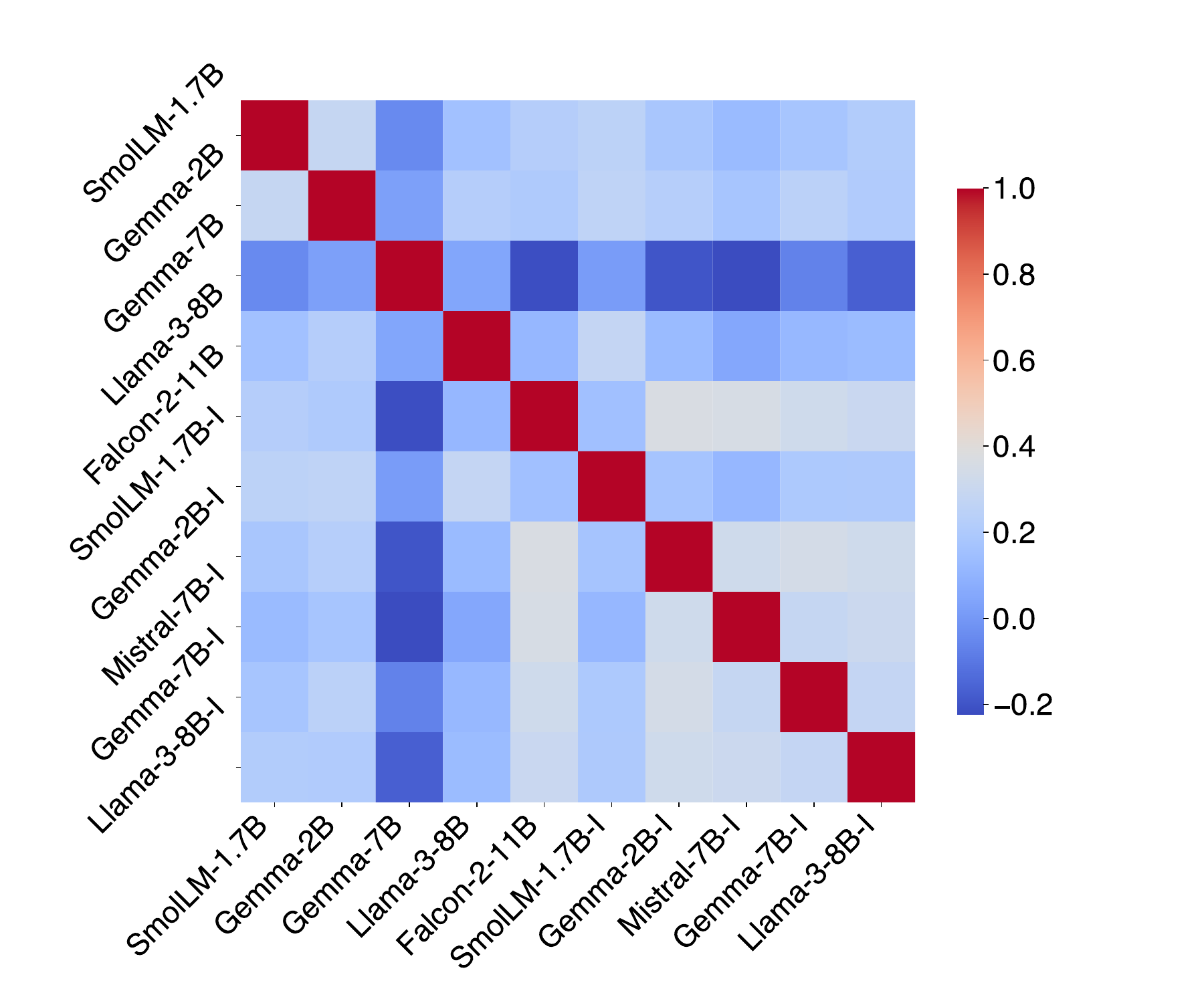}
    \caption{Correlation matrix of mean BERTScore recalls across different task instances for outputs of LMs.}
    \label{fig:correlation_lm}
\end{figure}

One of the hypothesis was that different LMs would perform differently. To demonstrate that, we show the correlation between BERTScore recalls of LM outputs, shown in Figure~\ref{fig:correlation_lm}, is low. This shows that their performance with different task types are inherently different, and therefore, selecting the right LM for a usage requirement becomes crucial. To analyze this, we detail their performance in our proposed evaluation framework. For these analyses, we use the best prompt style for that entity of that aspect (refer Appendix~\ref{app:prompt_line_graphs} to determine that).

\subsection{Comparison Across Task Types}
\label{sec:comparison_task_types}

\begin{figure*}[!ht]
    \centering
        \begin{subfigure}[b]{0.31\linewidth}
        \includegraphics[width=\linewidth]{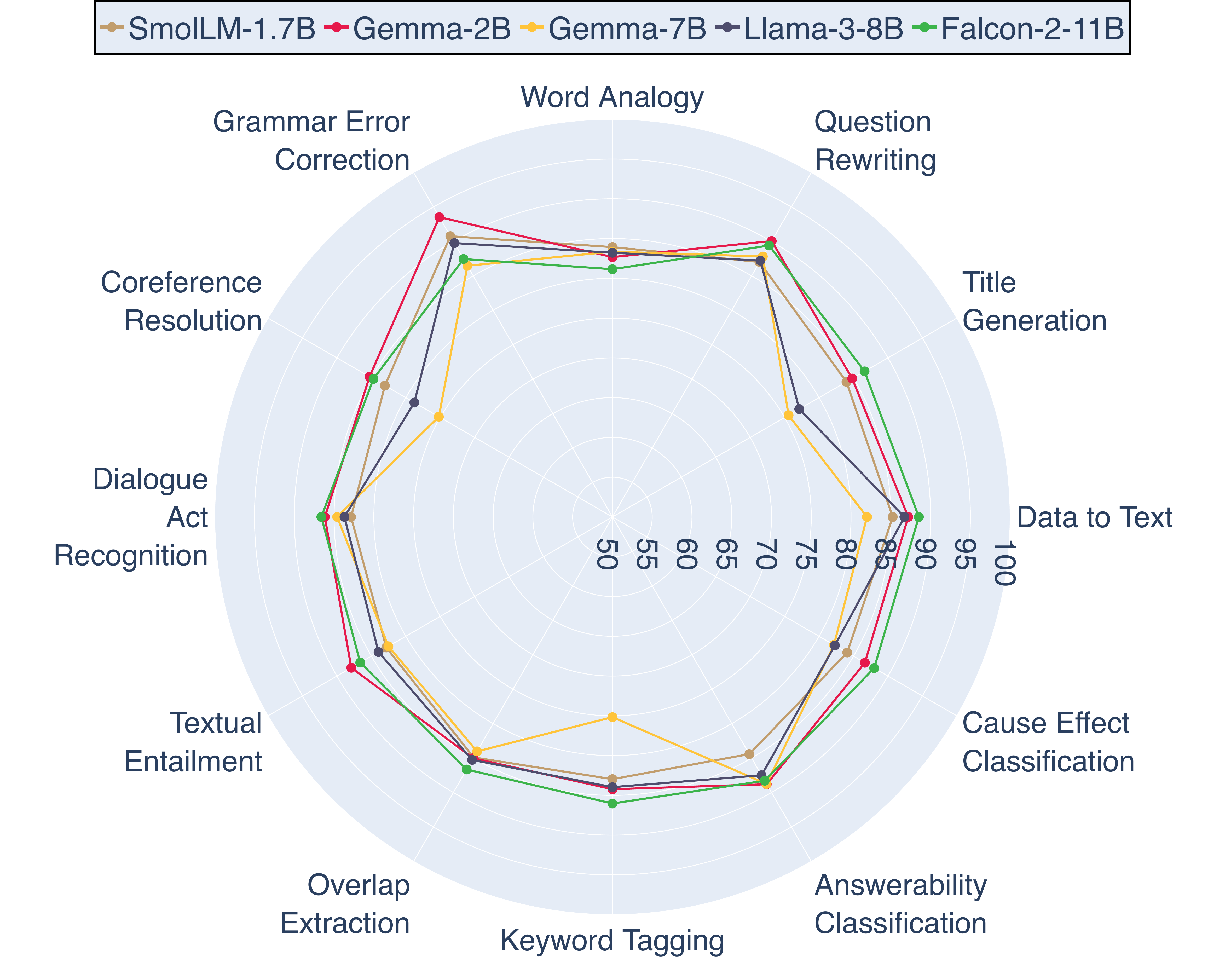}
        \caption{Task Types; Pre-trained Models}
        \label{fig:spider_chart_task_types_pt_models}
    \end{subfigure}
    \hfill
    \begin{subfigure}[b]{0.31\linewidth}
        \includegraphics[width=\linewidth]{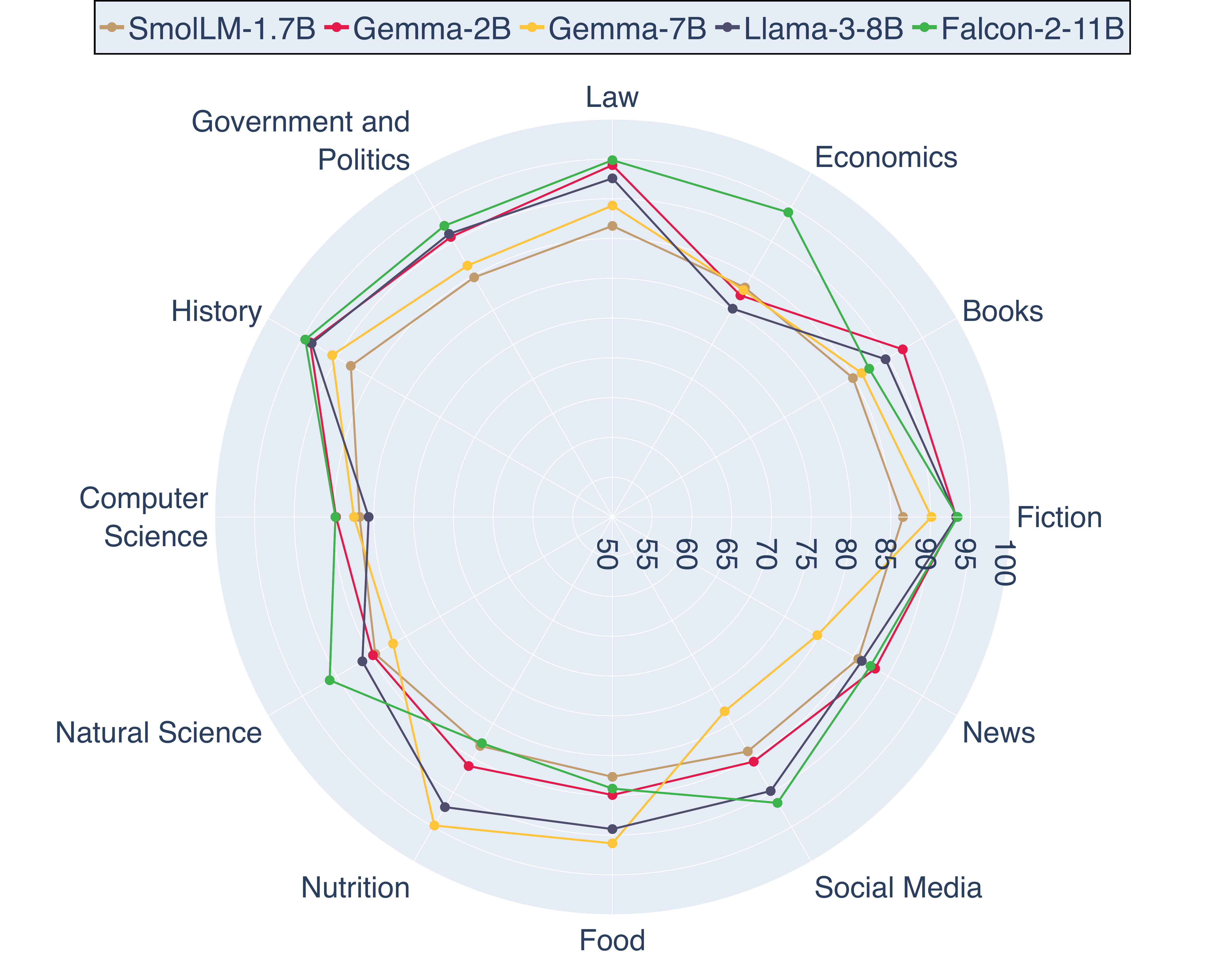}
        \caption{Domains; Pre-trained Models}
        \label{fig:spider_chart_domains_pt_models}
    \end{subfigure}
    \hfill
    \begin{subfigure}[b]{0.31\linewidth}
        \includegraphics[width=\linewidth]{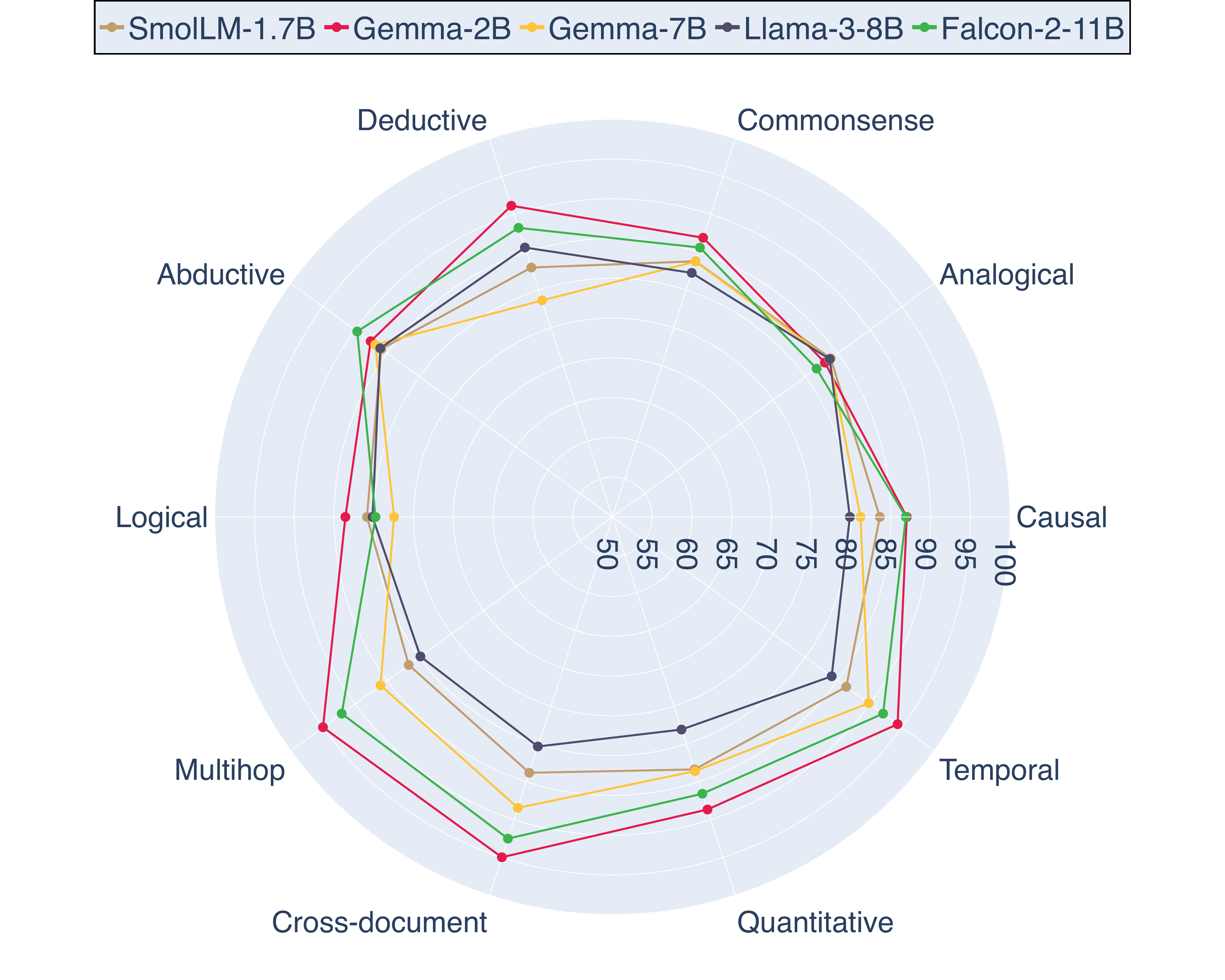}
        \caption{Reasonings; Pre-trained Models}
        \label{fig:spider_chart_reasoning_pt_models}
    \end{subfigure}

    \begin{subfigure}[b]{0.31\linewidth}
        \includegraphics[width=\linewidth]{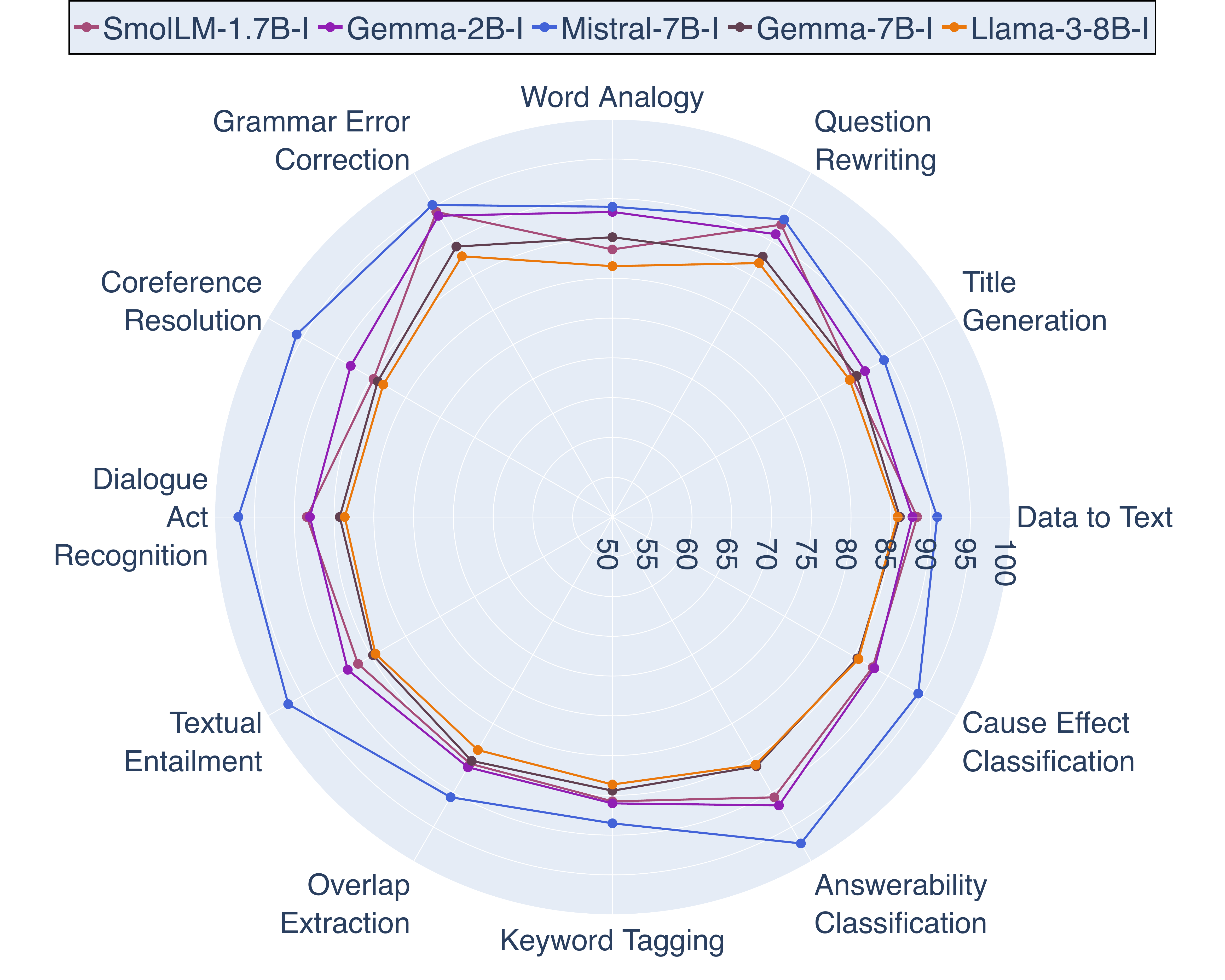}
        \caption{Task Types; IT Models}
        \label{fig:spider_chart_task_types_it_models}
    \end{subfigure}
    \hfill
    \begin{subfigure}[b]{0.31\linewidth}
        \includegraphics[width=\linewidth]{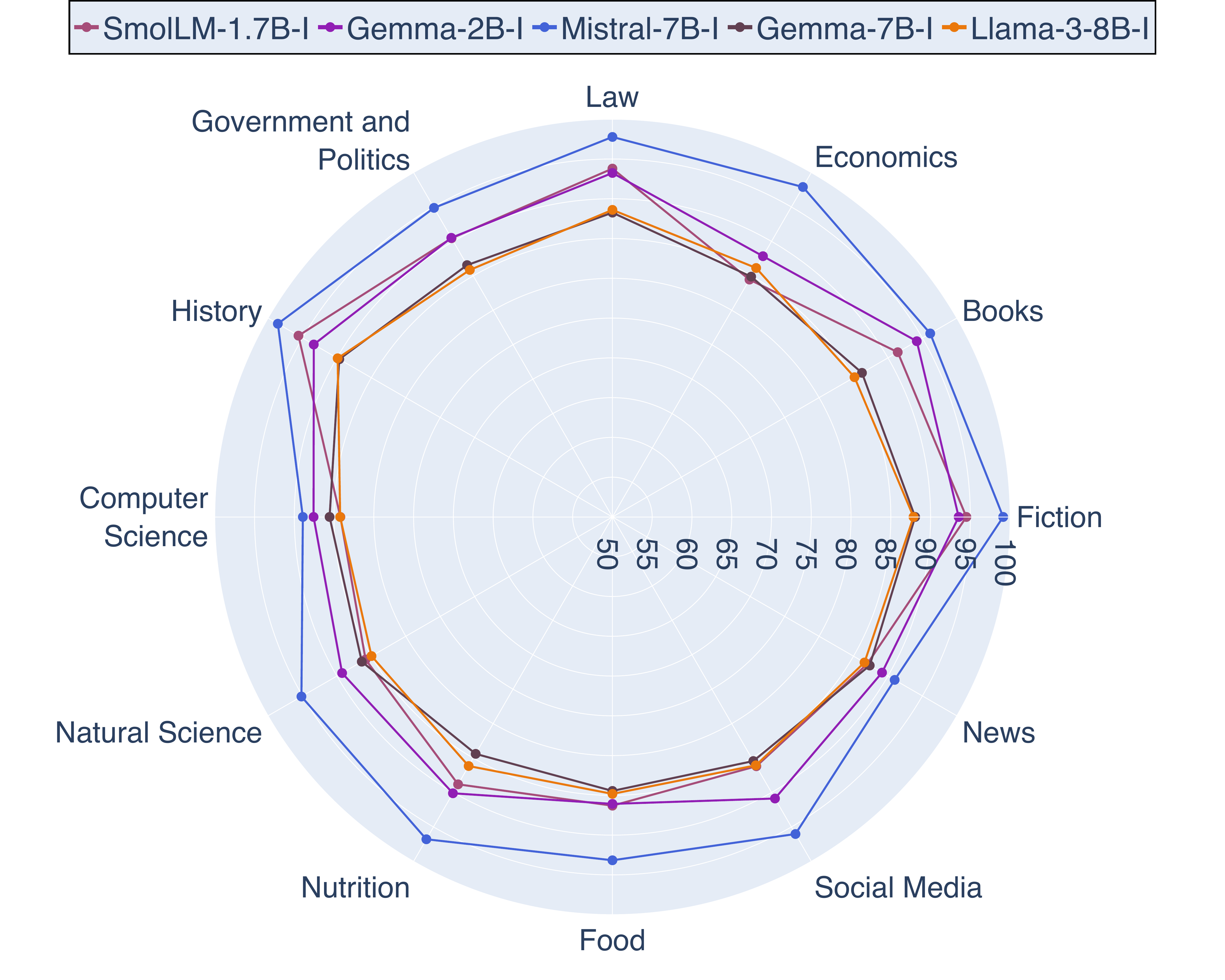}
        \caption{Domains; IT Models}
        \label{fig:spider_chart_domains_it_models}
    \end{subfigure}
    \hfill
    \begin{subfigure}[b]{0.31\linewidth}
        \includegraphics[width=\linewidth]{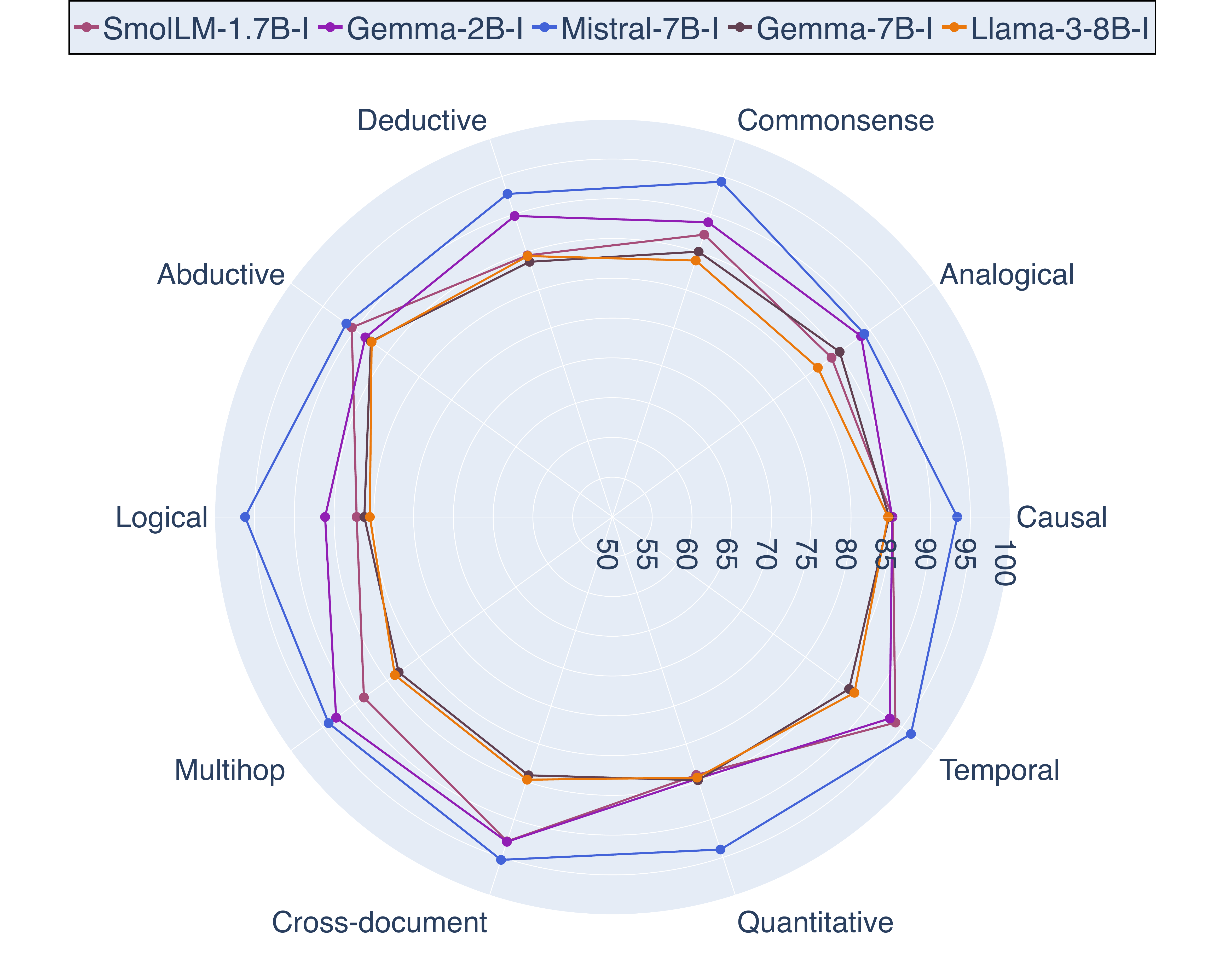}
        \caption{Reasonings; IT Models}
        \label{fig:spider_chart_reasoning_it_models}
    \end{subfigure}
    
    \caption{Mean BERTScore recall across various task types, domains, and reasoning types, segmented by pre-trained vs. instruction-tuned models (Note that the range doesn't start from 0 for better visibility).}
    \label{fig:aspect_result_variation_spider_charts}
\end{figure*}

Figure~\ref{fig:spider_chart_task_types_pt_models} and Figure~\ref{fig:spider_chart_task_types_it_models} show variation of performance on task-types for pre-trained and IT models. 

Most of the pre-trained models perform reasonably well on most tasks. We see that Gemma-2B always and SmolLM-1.7B sometimes perform better than all 7B and 8B models, which is opposite to the general understanding that scale improves performance. So, other design factors are also relevant which contribute to their strengths. Gemma-2B is the best across 50\% of the task types, with Falcon-2-11B leading in the remaining, except Word Analogy where SmolLM-1.7B is marginally the best. Considering the scale of the two models, Gemma-2B is a strong choice with resource constraints across all task types, unless Falcon-2-11B is needed purely on performance. Gemma-7B and Llama-3-8B hover below the top two with varying differences. We don't identify any patterns at group levels here but the difference between the top two models is similar across most tasks.

In IT models, Mistral-7B-I performs best on all task types, with Gemma-2B-I and SmolLM-1.7B-I competing for the second-best. At group level, we find the difference to be smaller for linguistic relationship and generation tasks, but large for semantic \& pragmatic analysis tasks. Like their pre-trained variants, Gemma-7B-I and Llama-3-8B-I seldom compete with Gemma-2B-I in some tasks, but never outperform it. So, Gemma-2B, SmolLM-1.7B-I and Mistral-7B-I can be selected based on performance and resources trade-offs.

\subsection{Comparison Across Application Domains}
\label{sec:comparison_domains}

The behavior of LMs across application domains can be visualized in Figure~\ref{fig:spider_chart_domains_pt_models} and \ref{fig:spider_chart_domains_it_models} for pre-trained and IT models, respectively.

Particularly for pre-trained models, the performance is very sensitive across domains. For social sciences \& humanities, and science \& technology domain groups, Falcon-2-11B performs the best with Gemma-2B and Llama-3-8B following. Gemma-2B and Falcon-2-11B are not always the best ones. In health and medical tasks, Gemma-7B outperforms all models. Falcon-2-11B and Gemma-2B suffer a significant performance degradation in this group. Therefore, for domains, the choice of pre-trained LMs depends on the use case and other constraints. SmolLM-1.7B felt like a strong choice in task types, but here we see that it struggles with these domains. Its strength in Section~\ref{sec:comparison_task_types} might be from other domains not considered here, showing its sensitivity with domains.

Among the IT models, we see similar trends as in task types - Mistral-7B-I being the best in all domains, and Gemma-2B-I and SmolLM-1.7B-I competing for second. The difference with Gemma-2B-I is closer in some domains like Computer Science, News, and Books, and largest in Economics. We also see that SmolLM-1.7B-I has strong limitations in Science and Technology group. Hence, Mistral-7B-I is still the best choice with best prompt style if the available resource allows, and if not, then Gemma-2B-I or SmolLM-1.7B is the way to proceed based on requirements.

Group-level behavior is more prominent in this aspect, highlighting the importance of our three-tier framework. Even in case of analyzing a new domain that is not present here, the performance of the group that domain would belong to can give an idea of baseline performance.

\subsection{Comparison Across Reasoning Types}
\label{sec:comparison_reasoning}

52 out of 119 task definitions in the dataset don't have a reasoning type as not all tasks require reasoning. For the remaining, the performance of different pre-trained LMs are shown in Figure~\ref{fig:spider_chart_reasoning_pt_models} and for all IT models in Figure~\ref{fig:spider_chart_reasoning_it_models}.

In the pre-trained models, we find that where reasoning is involved, Gemma-2B marginally outperforms Falcon-2-11B in all types of reasoning except Abductive reasoning, where it comes second by a small margin. It shows that Gemma-2B is a great choice where reasoning is involved, having advantage in both performance and model size. Llama-3-8B proves to be the best in analogical reasoning. In general, it is observed that the performance of all pre-trained LMs is the least for Comparative and Relational reasoning types, highlighting a potential common limitation of ability in these types of task in zero-shot. Therefore, adapting the LMs might become crucial in this case.

With IT models, behavior remains similar to the previous two aspects for all the five models, with Mistral-7B-I coming out to be a clear choice. The difference between Mistral-7B-I and Gemma-2B-I is minimum in complex inference \& analysis types, and maximum for types like logical and quantitative reasoning. SmolLM-1.7B-I also depicts weaknesses in some reasoning types. This shows that while choosing a pre-trained model has its complexities, for IT models, the choice is relatively simpler after considering external constraints.

The quantified performance of each entity of all three aspects in the dataset (even ones not included in Fig~\ref{fig:categories}) with each LM is given in Appendix~\ref{app:aspects_lm_performance}.

\subsection{Comparison with State-of-the-art LLMs}
\label{sec:comparison_sota}

\begin{figure*}[!ht]
    \centering
    \begin{subfigure}[b]{0.31\linewidth}
        \includegraphics[width=\linewidth]{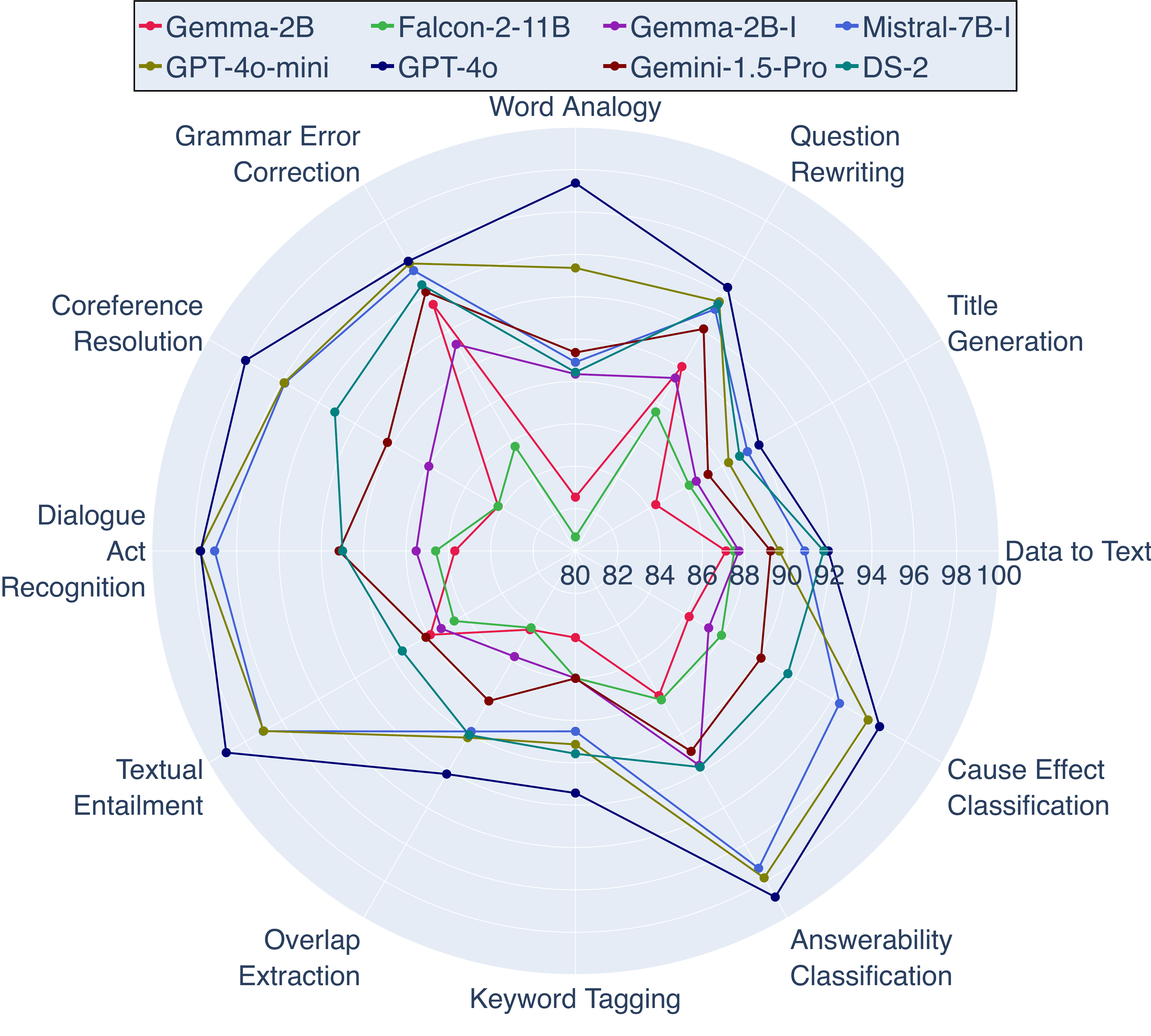}
        \caption{Task Types}
        \label{fig:spider_chart_task_types_sota_models}
    \end{subfigure}
    \hfill
    \begin{subfigure}[b]{0.31\linewidth}
        \includegraphics[width=\linewidth]{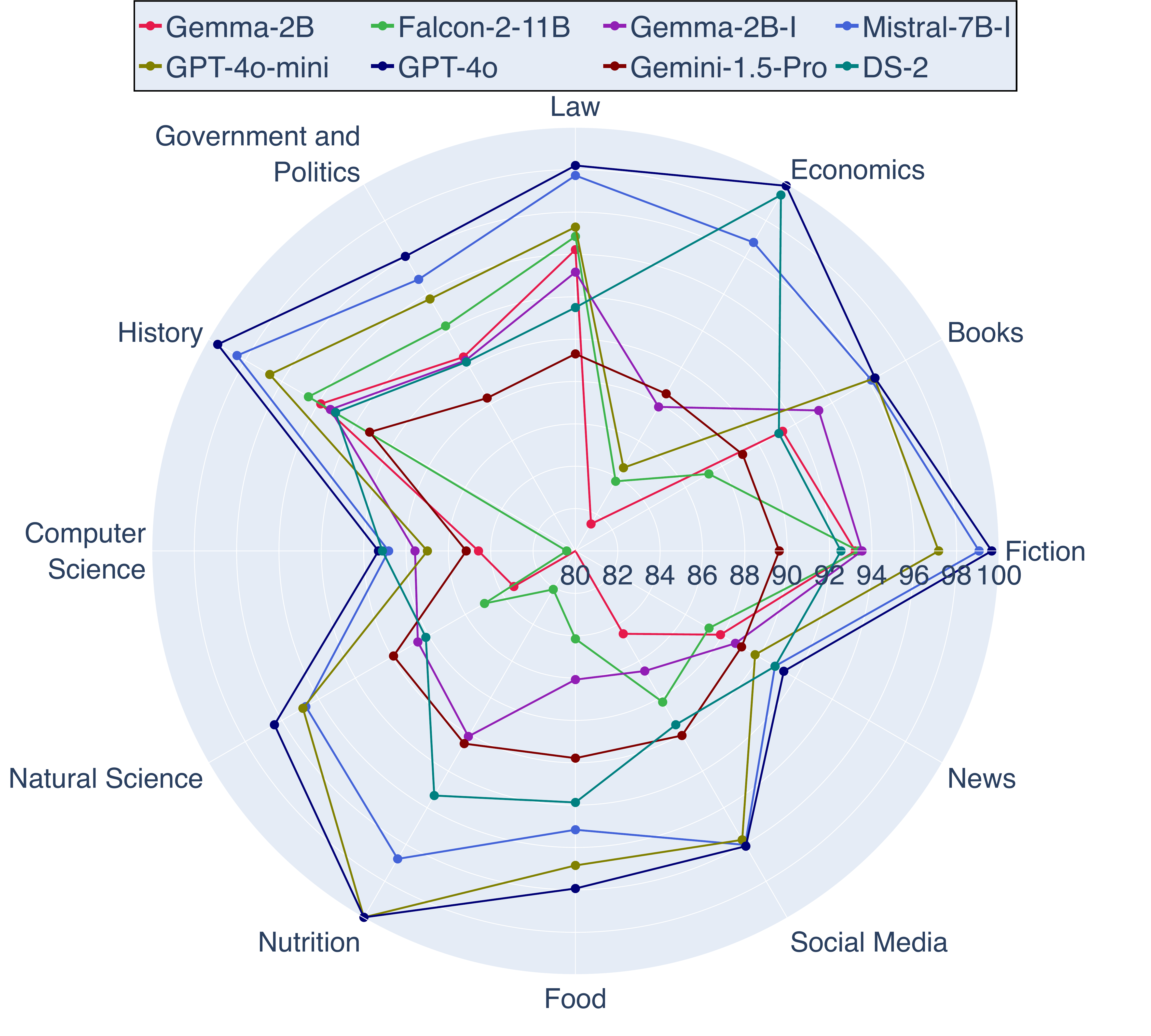}
        \caption{Domains}
        \label{fig:spider_chart_domains_sota_models}
    \end{subfigure}
    \hfill
    \begin{subfigure}[b]{0.31\linewidth}
        \includegraphics[width=\linewidth]{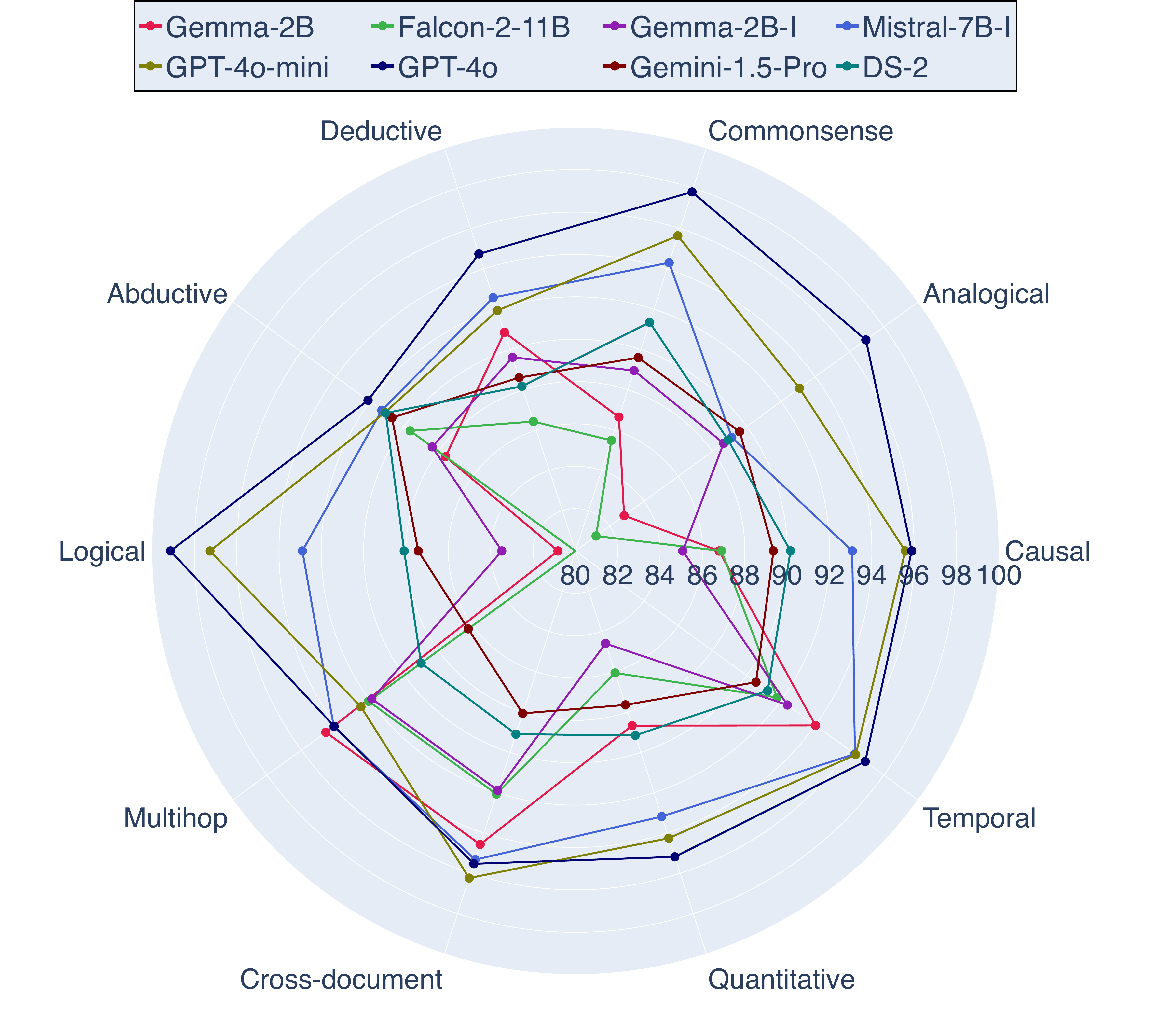}
        \caption{Reasonings}
        \label{fig:spider_chart_reasoning_sota_models}
    \end{subfigure}
    
    \caption{Mean BERTScore recall across various task types, domains, and reasoning types, compared against SOTA (Note the change in range and interval gaps for better visibility of small differences).}
    \label{fig:aspect_result_variation_spider_charts_sota}
\end{figure*}

We compare small, open LMs Gemma-2B, Falcon-11B, Mistral-7B-I and Gemma-2B-I (overall best two from each category) with recent SOTA LLMs like GPT-4o-mini, GPT-4o~\citep{gpt4,gpt4o}, and Gemini-1.5-Pro~\citep{gemini-1.5}. GPT-4o, Gemini-1.5-Pro and GPT-4o-mini are costly, large, closed models accessible using APIs. We use 8 examples with task definition for SOTA models, and report results in Figure~\ref{fig:aspect_result_variation_spider_charts_sota}.

We witness that Mistral-7B-I matches closely with all SOTA models globally. It's even very close to GPT-4o in some groups like Generation tasks, Art and Literature, and Media and Entertainment domains. All the 4 models outperform GPT-4o-mini, Gemini-1.5-Pro and DS-2 in many categories where they are strong, proving them to be a very strong choice. In application domains like in Social Sciences and Humanities group and Art and Literature group, Gemma-2B and Gemma-2B-I outperform Gemini-1.5-Pro as well. Being the open-sourced variant of a close family, this is commendable and shows that open LMs can be better choices than large or expensive ones in some usage scenarios. Many inferences can be drawn from the graph based a reader's need through this evaluation framework. From the average global \% decrease in performance reported in Table~\ref{tab:bert_score_decrease}, these models are globally competitive with the SOTA LLMs, proving their readiness in being utilized for practical applications with their other advantages as discussed previously. The gaps of pre-trained models are higher than IT models, but aligning them further for specific use can improve results. We also evaluate these SOTA LMs for all entities of each aspect in Appendix~\ref{app:aspects_lm_performance}.

\begin{table}[htbp]
    \centering
    \begin{tabular}{l|ccc}
        \toprule
        \textbf{LM} & \textbf{Gem-1.5} & \textbf{G-4o-m} & \textbf{G-4o} \\ 
        \toprule
        Gemma-2B     & 3.28\%           & 8.12\%          & 9.78\%        \\
        Falcon-2-11B & 3.54\%           & 8.37\%          & 10.02\%       \\ \midrule
        Gemma-2B-I   & 1.44\%           & 6.38\%          & 8.07\%        \\ 
        Mistral-7B-I & -4.94\%          & 0.32\%          & 2.12\%        \\
        \bottomrule
    \end{tabular}
    \caption{Avg. Percentage decrease in mean BERTScore recall of open LMs compared to Gemini-1.5-Pro (Gem-1.5), GPT-4o-mini (G-4o-m) and GPT-4o (G-4o).}
    \label{tab:bert_score_decrease}
\end{table}

\subsection{Comparison Across Prompt Styles}
\label{sec:comparison_prompts}

Language models' behavior depends significantly on the prompts. Writing good task descriptions and in-context examples requires time, good understanding of subtle variations, sufficient domain knowledge, etc., which is not straightforward. So, we analyze how the performance varies for each entity of each aspect with changing instruction, focusing on the best performing IT model - Mistral-7B-I, since it can directly be used if prompted correctly.

\begin{figure*}[!ht]
    \centering
    \begin{subfigure}[b]{0.31\linewidth}
        \includegraphics[width=\linewidth]{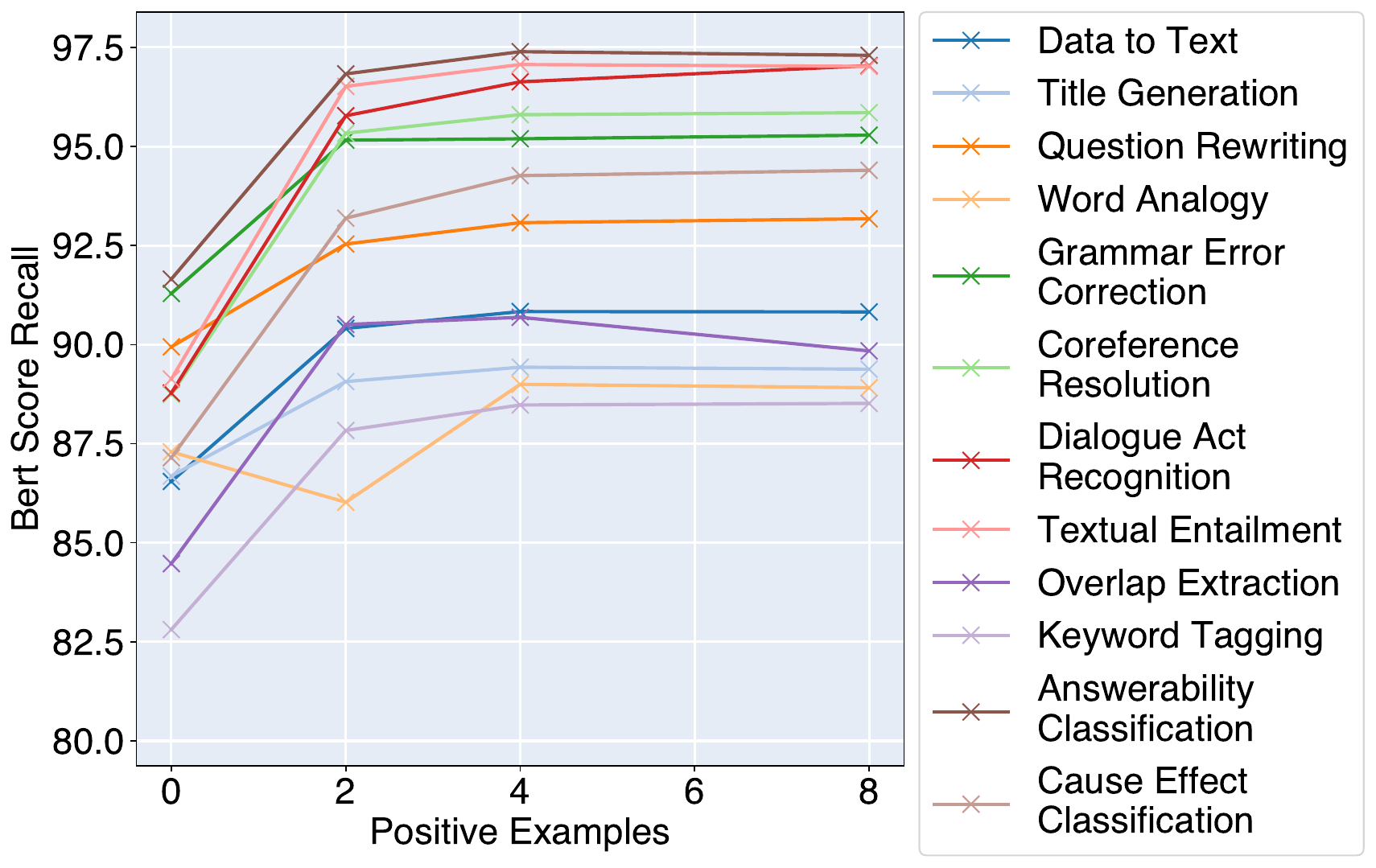}
        \caption{Task Types; with definition}
        \label{fig:line_graph_task_types_inst_with_def_mistral}
    \end{subfigure}
    \hfill
    \begin{subfigure}[b]{0.31\linewidth}
        \includegraphics[width=\linewidth]{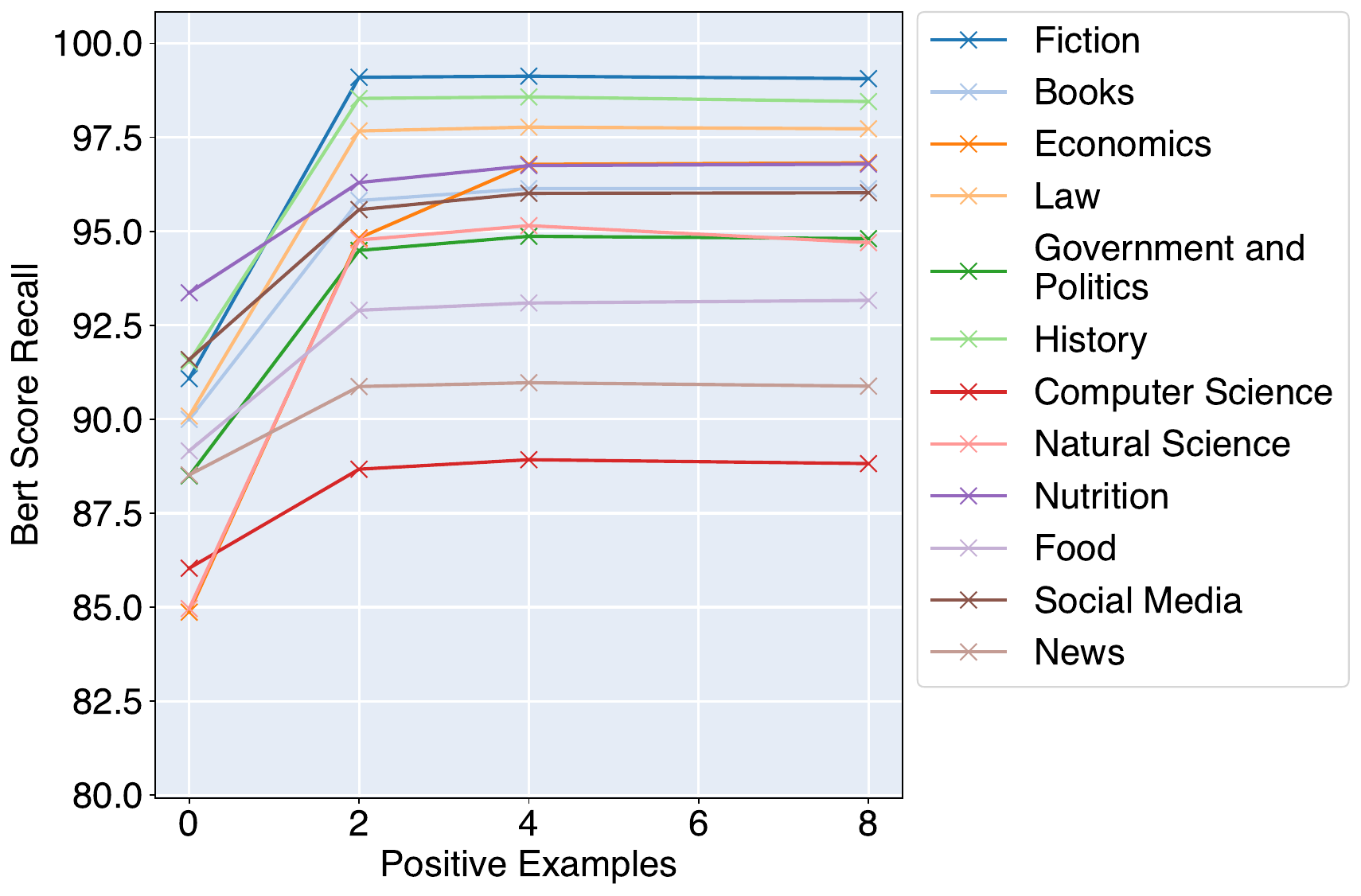}
        \caption{Domains; with definition}
        \label{fig:line_graph_domains_inst_with_def_mistral}
    \end{subfigure}
    \hfill
    \begin{subfigure}[b]{0.31\linewidth}
        \includegraphics[width=\linewidth]{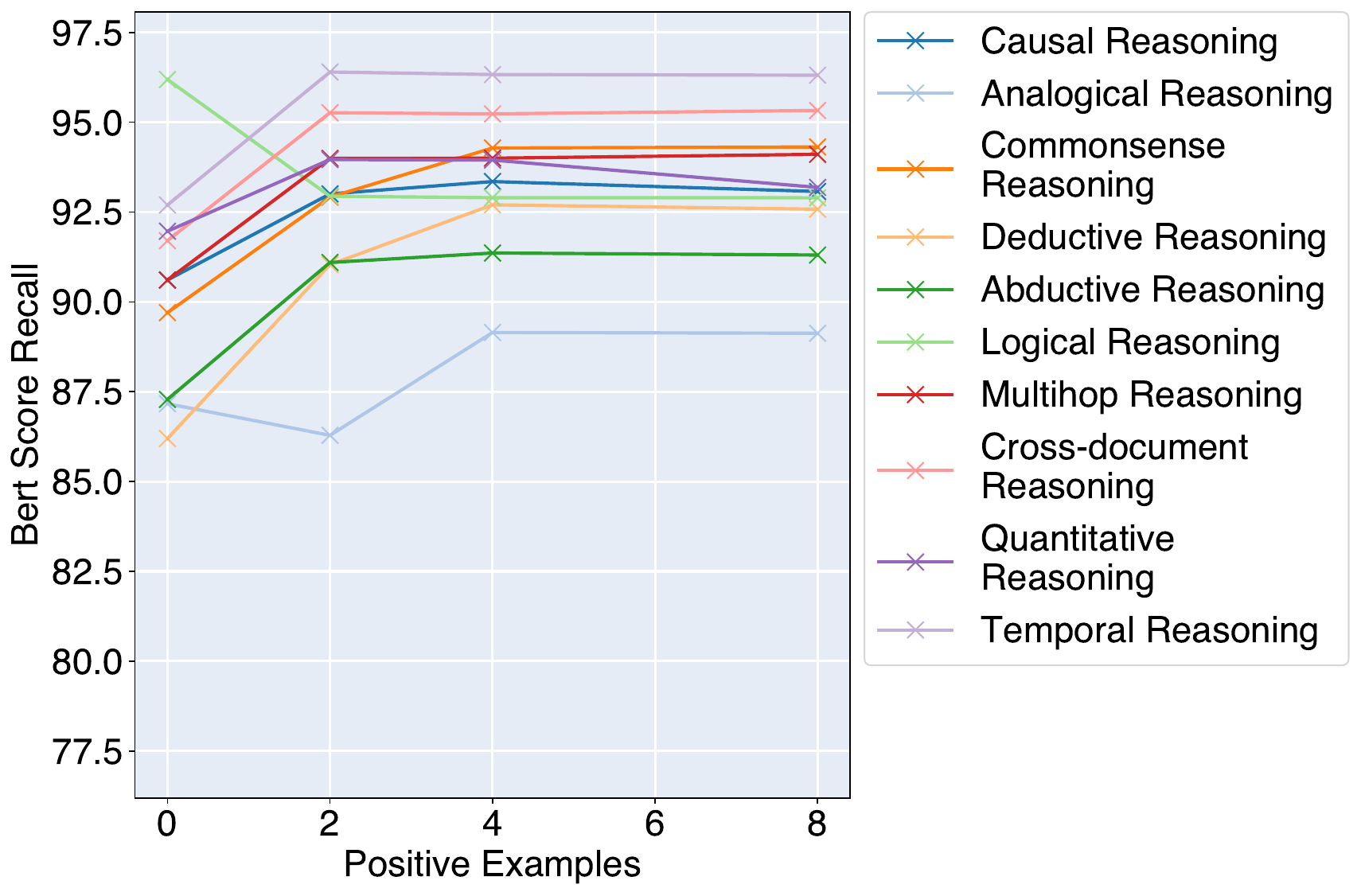}
        \caption{Reasonings; with definition}
        \label{fig:line_graph_reasoning_inst_with_def_mistral}
    \end{subfigure}

    \begin{subfigure}[b]{0.31\linewidth}
        \includegraphics[width=\linewidth]{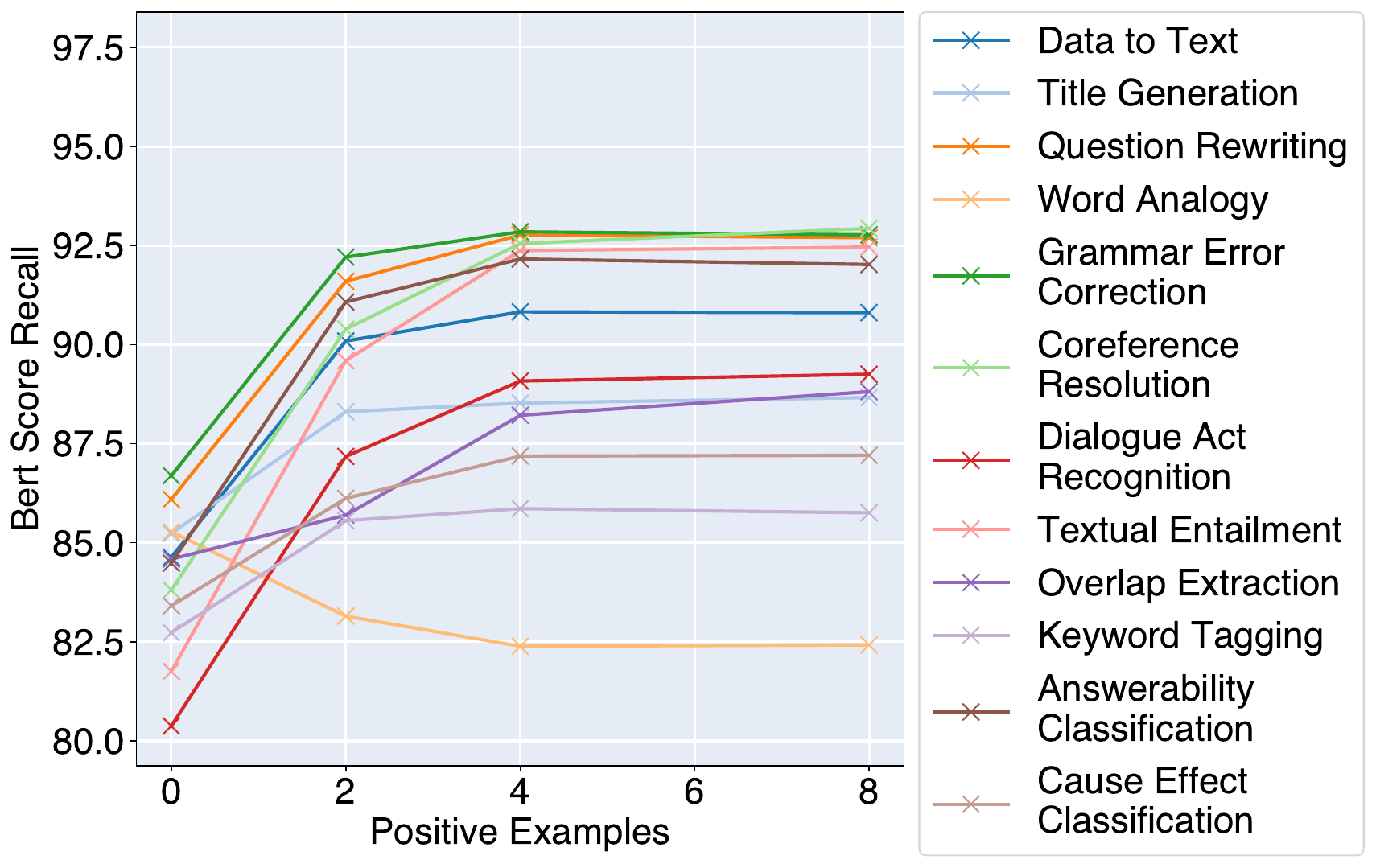}
        \caption{Task Types; without definition}
        \label{fig:line_graph_task_types_inst_without_def_mistral}
    \end{subfigure}
    \hfill
    \begin{subfigure}[b]{0.31\linewidth}
        \includegraphics[width=\linewidth]{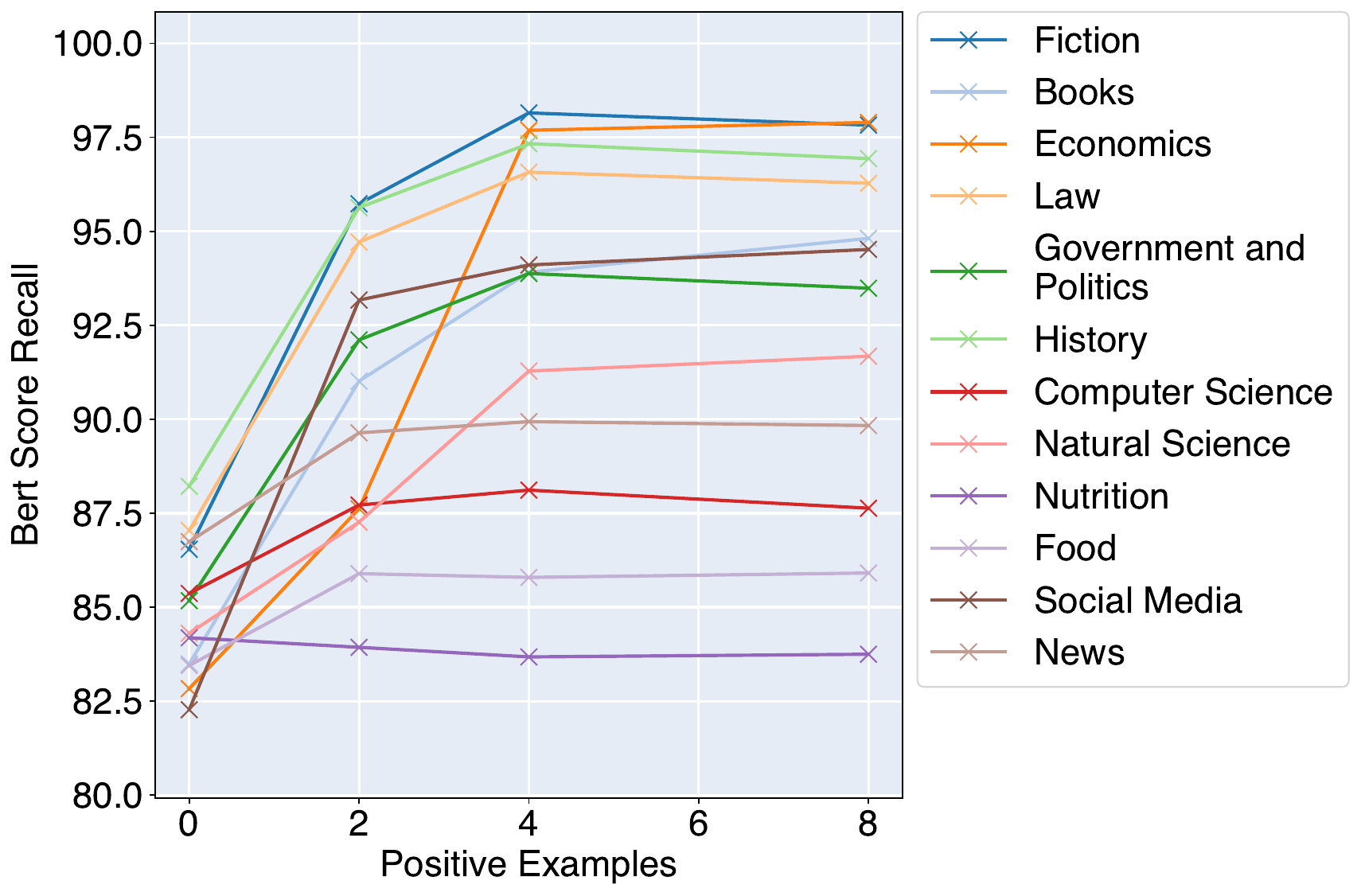}
        \caption{Domains; without definition}
        \label{fig:line_graph_domains_inst_without_def_mistral}
    \end{subfigure}
    \hfill
    \begin{subfigure}[b]{0.31\linewidth}
        \includegraphics[width=\linewidth]{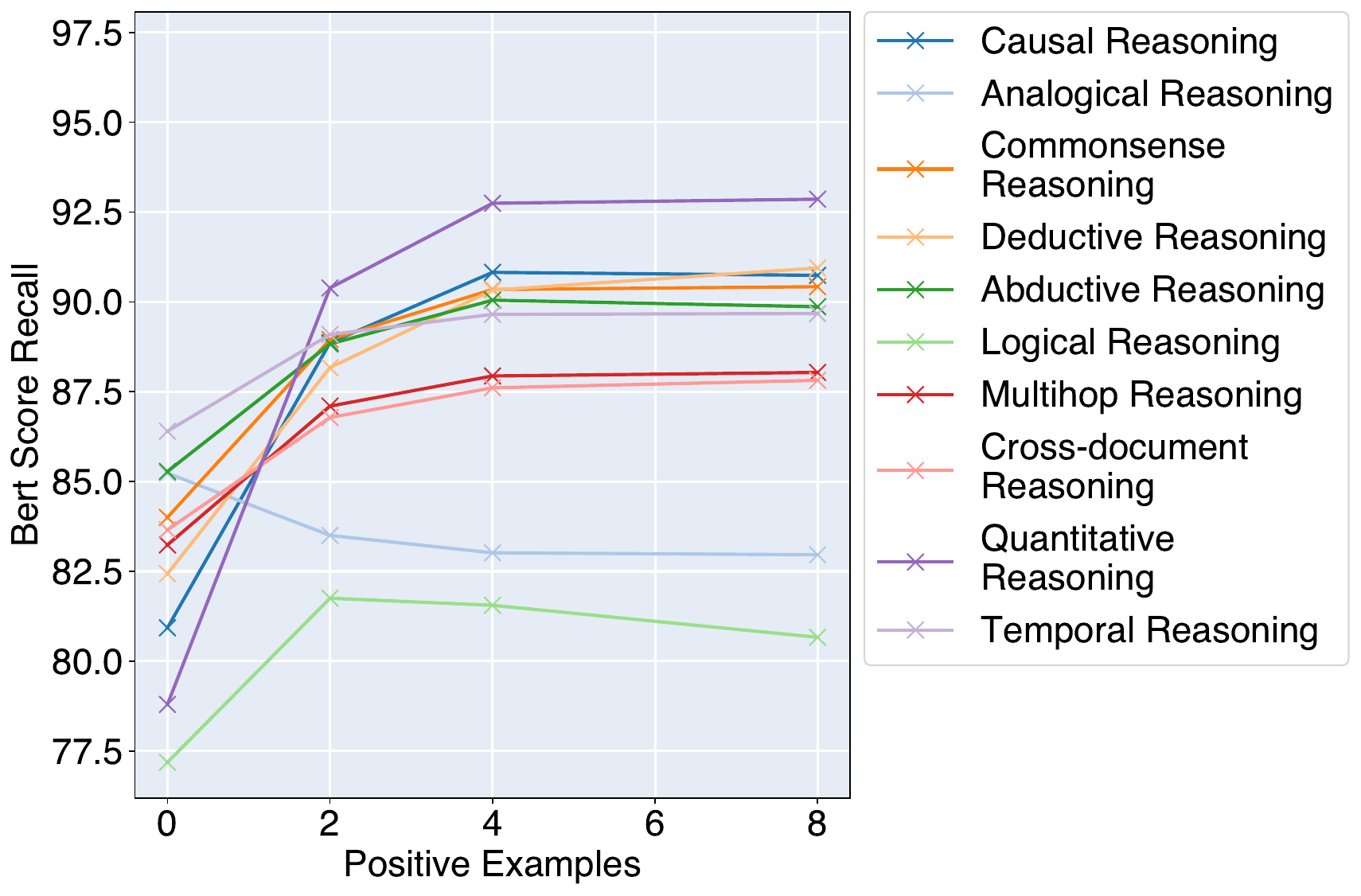}
        \caption{Reasonings; without definition}
        \label{fig:line_graph_reasoning_inst_without_def_mistral}
    \end{subfigure}
    
    \caption{Mean BERTScore recall for Mistral-7B-I for task types, domains, and reasoning types by varying in-context examples, segmented by with and without using task definitions.}
    \label{fig:instruction_variation_Mistral-7B-I}
\end{figure*}

We visualize the results in Figure~\ref{fig:instruction_variation_Mistral-7B-I}. Using this, users can analyze the trade-offs of crafting instructions versus its possible impact on performance.

On initial analysis, using chat-style definitions proves better, but the performance increase looks small after 2 examples. So, using 2 examples can suffice. This trend is consistent for most entities across all three aspects. However, adding definition impacts different entities differently. For example, dialogue act recognition's performance on zero examples increases from 80.37 to 88.77 just by including task definition. But, for keyword tagging, the change is from 82.73 to only 82.81. We also see behaviors like Word Analogy, for which more examples negatively impact the output if definition is not provided. It may be because in absence of clear instruction, the model fails to comprehend the task from examples. Further, taking `Social Media', adding task definition increases performance from 82.27 to 91.58 without examples, but, adding 2 examples without definition also improves score to 93.17. So, a choice is available between definition and examples. The rate of improvement with adding examples is also different for different entities. Some tasks don't have 8 examples in the dataset, so 4 to 8 example transition should be inferred accordingly.

Using these graphs, one can determine a prompt style for an application within other constraints of ability, cost, need, etc. in crafting instructions. These trends are different for different LMs. So, we have included these line graphs for all other LMs in Appendix~\ref{app:prompt_line_graphs}. This will also help in analyzing best prompt style and studying relative performance difference of each entity of each aspect.

\subsection{Task Definition v/s Paraphrased Definition}
\label{sec:comparison_paraphrased_def}

To evaluate dependency of models to the provided task definition, we also evaluate them with their paraphrases. These are generated using \texttt{gpt-3.5-turbo}~\citep{gpt3, gpt35turbo}, and used with best in-context example count as per Table~\ref{tab:all_metrics_results}. Then, results are evaluated using the same pipeline, and reported in Table~\ref{tab:paraphrased_def_results} for the two-best performing LMs in each category.

\begin{table}[ht]
    \centering
    \begin{tabular}{l|c|cc}
        \toprule
        \textbf{Model Name} & \textbf{Ex.} & \textbf{Def} & \textbf{Par. Def.} \\ 
        \toprule
        Gemma-2B & 4 & 86.41 & 85.77 \\
        Falcon-2-11B & 8 & 86.18 & 86.00 \\ 
        \midrule
        Gemma-2B-I & 4 & 87.96 & 87.67 \\
        Mistral-7B-I & 8 & 93.76 & 93.22 \\ 
        \bottomrule
    \end{tabular}
    \caption{Mean BERTScore recall values of outputs with actual task definition (Def) versus paraphrased definitions (Par. Def) using 'Ex.' in-context examples.}
    \label{tab:paraphrased_def_results}
\end{table}

The median decrease in performance for all 10 LMs also is only 0.35\%., which can be attributed to some loss of information during paraphrasing. But, most of the models prove robust to perturbations in task definitions, as long as a prompt can reasonably explain the task. Appendix~\ref{app:paraphrasing_definitions} has more details on obtaining paraphrases and results on all LMs.

\section{High-level Takeaways}

We find that recent, open and small-scale Language Models (LMs) are very effective. Detailed recommendations on LMs and their performance trends in different groups and entities are discussed in depth in Sections~\ref{sec:comparison_task_types},~\ref{sec:comparison_domains} and~\ref{sec:comparison_reasoning}, but we summarize them in the below paragraphs too. Although it is visible that no single LM is a global solution, but, if selected and used appropriately with an effective prompt style for a task type, domain, or reasoning type, they can perform within 10\% (worst case) of SOTA LLMs like GPT-4o, and outperform DS-2, GPT-4o-mini, and Gemini-1.5-Pro with advantages in efficiency, control and cost.

For the LMs we experimented with, among pre-trained models, we recommend using Gemma-2B and Falcon-2-11B based on different aspects and entities, but sometimes, Gemma-7B, Llama-3-8B can be great choices (also detailed in Appendix~\ref{app:aspects_lm_performance}). The performance of pre-trained models can be taken as a measure of their knowledge of different use-cases. Based on other factors like availability, compliance, size, right LM can be selected and customized as needed. Limitations of some pre-trained models are discussed in Appendix~\ref{app:pt_model_limitations}.

For IT models, Mistral-7B-I is a clear best in all aspects, and Gemma-2B-I and SmolLM-1.7B-I come second in most cases. Since these models are IT, they can be used directly with chat-style description and examples. We recommend a model in these three (and other models), based on other factors like size, licensing, etc. Some qualitative outputs of Mistral-7B-I are given in Appendix~\ref{app:qual_examples}.

We also study the performance trade-off for multiple prompt styles and recommend the best one for these models. As noted in Section~\ref{sec:comparison_prompts}, having a chat-style task description to guide the LM is recommended. But, having more in-context examples is not always better, and considering use-case and LMs, the right number of example can vary. The models are also robust to changes in task definitions, if it can provide all (or most) information to complete the task. They are also reasonably robust to subtle intended/unintended incorrectness in definitions, which is analyzed in Appendix~\ref{app:adversarial_definitions}. In appendix~\ref{app:aspects_lm_performance}, we also compare and show that the small LMs even outperform DeepSeek-v2 (DS-2)~\citep{deepseekv2} in many categories.
\section{Conclusion}

We identify some limitations of using SOTA, proprietary LLMs and show that open LMs with 1.7B--11B parameters can be effective for applications. We create a three-tier evaluation framework and analyze semantic correctness of output of 10 LMs across multiple hierarchical umbrellas. Using this framework, we demonstrate that while these models don't work best in every scenario, if selected properly, they are effective and can compete with and outperform models like Gemini-1.5-Pro, GPT-4o-mini and GPT-4o despite being 10-20 folds smaller in size. We also provide a guide in Appendix~\ref{app:guide} on how one can this work to select an LM for one's specific needs. We hope that our contributions will enable the community to make a confident shift towards considering using these small, open LMs for their need.
\section{Limitations}
\label{sec:limitations}

\textbf{Limitations of Dataset}: We derive our experimental dataset from the test-set of Super Natural Instructions~\citep{super_natural_instructions} and take the labels of aspects from there itself. We therefore assume that they are labeled correctly for task types, domains and reasoning types. There are many other task types, domains and reasoning types not available in its test set, which we were not able to consider. However, if an entity of an aspect is not present, one can leverage the performance of the groups that we created in Figure~\ref{fig:categories}, or choose the nearest entity (from Section~\ref{sec:exp_results}/ Appendix~\ref{app:aspects_lm_performance}) and roughly estimate the performance. We were also bounded in prompts by the examples and task definitions available. But, we did experiment by paraphrasing definitions in Section~\ref{sec:comparison_paraphrased_def}, Appendix~\ref{app:paraphrasing_definitions} to limit this to some extent. Using this dataset however may not bring significant dataset bias, as this is a meta-dataset curated using several NLP benchmark datasets.

\textbf{Limitations of LMs}: There are many LMs proposed by the research community, and it is not practically feasible to experiment with all of them. Further, the definition of a small LM is also relative. We selected the LMs based on the ones which have come out recently and promise strong capabilities. Although we capped our LMs at 11B parameters, we didn't find the performance to be a direct function of number of parameters, so we believe this decision should not have too drastic effects. We also didn't evaluate biases~\citep{bias_fairness_survey} and other factors other than semantic correctness of generated results of these models, but the models' technical reports~\citep{smollm,gemma,llama3,falcon,falcon2} provide more details on those.

\textbf{Limitations of Prompts}: We experimented with 8 prompt styles, apart from using paraphrased definitions, adversarial definitions. But, all of them (excluding paraphrased and adversarial definitions) were built using the elements of the dataset available. We acknowledge that there may be some tasks where another prompt style or using more domain-adapted prompts perform better. Additionally, if the LM is adapted/fine-tuned in any way, the best prompt style can change based on the data and technique used for it. However, to keep a standard and common features across tasks, we intentionally chose this approach. This study should provide an initial idea of whether descriptions are needed and the number of examples required when using the LM without any changes.

\textbf{Assumptions in Reporting of Results}: We are considering the impact of each aspect one at a time when reporting results. For example, in pre-trained models, we see that Gemma-2B is best for Grammar Error Correction, and Falcon-2-11B is best for Economics domain. But what if there is a task instance that involves grammar error correction for an Economics article? This can sometimes give a dual outcome, with one LM recommended for task type (grammar error correction), and one for domain (Economics). To eliminate this, we tried to do a pairwise aspect analysis, but in the dataset, 86.86\% of task type-domain and 88.25\% of domain-reasoning type pairs had no task instances. We could have generated labels of aspect entities using other techniques, or could've generated artificial data to fill these gaps, but we didn't want conflicting sources of experimental data as they could created additional undesired biases/variations of sources, type of data. Therefore, due to lack of sufficient labels, we didn't report those results. However, despite this independent assumption, this work can still help in narrowing down to 2-3 models which will be helpful. We also assume that the results reported by our experimental data represents the actual performance of that LM on that entity type. This may not be correct but considering Super Natural Instructions to be a meta-dataset of many other datasets, we believe it is a reasonable assumption. 

\section{Ethical Considerations}

This work evaluates performance of Language Models in terms of semantic correctness of outputs on various task types, application domains and reasoning types using different prompt styles. While we only included the entities that help the community, one can utilize/extrapolate the conclusions of this work for applications that are harmful. Further, one can create prompts using task definitions, in-context examples to extract negative behavior from the LMs, or attempt adversarial attacks on these LMs. We strictly discourage utilizing the results of this work or LMs in general in such ways. We also didn't evaluate these LMs on Bias and Fairness as it was out of scope of this paper. This work~\citep{bias_fairness_survey} discusses different types of biases and mitigation strategies.


\bibliography{anthology_2, custom}

\newpage
\appendix
\section*{Appendices}

\section{Guide to LM Selection for Your Application Using this Work}
\label{app:guide}

Before coming to this paper, finalize other constraints of your solution - resource availability, data availability, system constraints, economic parameters, expectation of results, etc. These are outside the scope of this work, but will help in choosing LMs based on this work. 

Then, check the relative performance of LMs for your task type/domain/reasoning type (or a combination). Find the closest available entity, and look up the performance of LMs of interest from Tables~\ref{tab:all_task_type_model_results},~\ref{tab:all_domain_model_results},~\ref{tab:all_reasoning_model_results}. From there, consider some options based on other constraints. For example, if you are planning to further align LMs on your task using any technique, choose from pre-trained models, if not, utilizing IT models will likely yeild better results. If you are bounded by resources, consider using smaller models that fit the requirements, or if you are bound by business/regulatory constraints, choose accordingly. 

Next, look-up those LMs and entities in Figure 8--17 to find the prompt style that gives best results. This will be less important if you are planning to fine-tune your LM or use a more domain-adapted prompt. But if not, this will be beneficial. Decide if you can use the best prompt style, and if not, what is the performance trade-off with styles you can use. Finalize the feasible prompt style from here.

Based on these selections and other design constraints, implement your solution.

This work is accompanied by a GitHub repository linked in the first page of the paper as a utility which will allow evaluating any LM as per this framework and generating visualizations. It supports evaluation and generation of visualizations on other evaluation metrics that are discussed in Table~\ref{tab:all_metrics_results}, and on a different set of task types, application domain and reasoning types as needed with minor configuration changes. No code change will be needed for utilizing HuggingFace implemented models. Usage guidelines are available in the README of the repository.

\section{Aspect-level Analyses}
\label{app:aspects_lm_performance}

In this appendix, we report results of all 14 LMs (5 pre-trained, 5 IT and 4 SOTA models that we compared our work to) on all entities of all three aspects present in the test set of the dataset. It includes the ones not covered in Section~\ref{sec:selection_of_aspects}, but were available in the test-set of Super-Natural Instructions~\citep{super_natural_instructions}, with English as the input and output languages. Note, we also provide the results SOTA models for comparisons. Table~\ref{tab:all_task_type_model_results} reports the results for all task types, Table~\ref{tab:all_domain_model_results} reports the results on all application domains and Table~\ref{tab:all_reasoning_model_results} for all reasoning types. Note, we abbreviate the model names at some places in the columns of these tables. The abbreviations and full model names can be found in Table~\ref{tab:abbreviation_table_models}.

In all our analyses, each domain has been considered independent, which is not always the case. There can be some tasks which can be classified into two aspects, like title generation for News articles will belong to title generation task type and News domain. However, in the dataset, there are many such pairwise aspects that do not contain any tasks, and for most of the ones that were present, Mistral-7B-I was the best model. Thus, we are not reporting the tabulated results for aspects considered pairwise considering the sparsity and repetitiveness of such a dense table. This is also discussed in Section~\ref{sec:limitations}. 
\begin{table}[htbp]
    \centering
    \begin{tabular}{c|c}
    \toprule
    \textbf{Abbreviation} & \textbf{Model name} \\
    \midrule
    S-1.7B & SmolLM-1.7B \\
    G-2B & Gemma-2B \\
    G-7B & Gemma-7B \\
    L-3-8B & Meta-Llama-3-8B \\
    F-2-11B & Falcon-2-11B \\
    S-1.7B-I & SmolLM-1.7B-I \\
    G-2B-I & Gemma-2B-I \\
    M-7B-I & Mistral-7B-I-v0.3 \\
    G-7B-I & Gemma-7B-I \\
    L-3-8B-I & Meta Llama-3-8B-I \\
    GPT-4o-m & GPT-4o-mini \\
    GPT-4o & GPT-4o \\
    DS-2 & DeepSeek-v2 \\
    Gem-1.5-Pro & Gemini-1.5-Pro \\
    \bottomrule
    \end{tabular}
    \caption{Abbreviation for model names.}
    \label{tab:abbreviation_table_models}
\end{table}

\begin{sidewaystable}
    \small
    \centering
    \begin{tabularx}{\textwidth}{
        >{\hsize=1.2\hsize}X | 
        >{\hsize=0.2\hsize}X | 
        >{\hsize=0.2\hsize}X  
        >{\hsize=0.2\hsize}X  
        >{\hsize=0.2\hsize}X  
        >{\hsize=0.2\hsize}X  
        >{\hsize=0.2\hsize}X | 
        >{\hsize=0.2\hsize}X  
        >{\hsize=0.2\hsize}X  
        >{\hsize=0.2\hsize}X  
        >{\hsize=0.2\hsize}X  
        >{\hsize=0.2\hsize}X | 
        >{\hsize=0.2\hsize}X  
        >{\hsize=0.2\hsize}X  
        >{\hsize=0.2\hsize}X  
        >{\hsize=0.2\hsize}X  
    }
        \toprule
        \textbf{Task Type} & \textbf{\# Inst.} & \multicolumn{5}{c}{\textbf{Pre-trained Models}} & \multicolumn{5}{c}{\textbf{Instruction-tuned Models}} & \multicolumn{4}{c}{\textbf{SOTA Models}} \\
        \cmidrule(lr){3-7} \cmidrule(lr){8-12} \cmidrule(lr){13-16}
        & & \textbf{S-1.7B} & \textbf{G-2B} & \textbf{G-7B} & \textbf{L-3-8B} & \textbf{F-2-11B} & \textbf{G-2B-I} & \textbf{S-1.7B-I} & \textbf{M-7B-I} & \textbf{G-7B-I} & \textbf{L-3-8B-I} & \textbf{GPT-4o-m} & \textbf{GPT-4o} & \textbf{DS-2} & \textbf{Gem-1.5-Pro} \\
        \midrule
        Answerability Classification & 1300 & 84.42 & \textbf{88.80} & 88.76 & 87.50 & 88.29 & 91.87 & 90.70 & \textbf{97.39} & 86.21 & 85.98 & 97.82 & 98.87 & 91.78 & 90.93 \\
        Cause Effect Classification & 700 & 84.09 & 86.65 & 82.16 & 82.30 & \textbf{88.00} & 88.03 & 87.79 & \textbf{94.40} & 85.56 & 85.73 & 95.96 & 96.59 & 91.58 & 90.12 \\
        Coreference Resolution & 1400 & 83.03 & \textbf{85.26} & 75.20 & 78.76 & 84.70 & 88.00 & 84.68 & \textbf{95.85} & 84.08 & 83.27 & 95.88 & 97.99 & 93.12 & 90.25 \\
        Data to Text & 826 & 85.25 & 87.19 & 82.01 & 86.71 & \textbf{88.53} & 87.72 & 88.30 & \textbf{90.83} & 86.14 & 85.87 & 91.55 & 89.62 & 91.73 & 89.22 \\
        Dialogue Act Recognition & 700 & 82.87 & 86.16 & 84.61 & 83.70 & \textbf{86.60} & 88.04 & 88.42 & \textbf{97.04} & 84.28 & 83.66 & 97.74 & 97.70 & 91.00 & 91.16 \\
        Grammar Error Correction & 100 & 90.75 & \textbf{93.52} & 86.47 & 89.76 & 87.45 & 93.72 & 94.29 & \textbf{95.29} & 89.25 & 87.84 & 95.67 & 95.79 & 94.51 & 94.13 \\
        Keyword Tagging & 500 & 82.97 & 84.23 & 75.17 & 83.97 & \textbf{86.03} & 86.01 & 85.75 & \textbf{88.52} & 84.42 & 83.64 & 89.13 & 91.43 & 89.57 & 86.01 \\
        Overlap Extraction & 200 & 84.91 & 85.03 & 84.03 & 85.27 & \textbf{86.66} & 86.34 & 85.82 & \textbf{90.69} & 85.40 & 83.83 & 90.18 & 92.16 & 90.04 & 88.18 \\
        Question Rewriting & 1100 & 87.01 & \textbf{90.05} & 87.81 & 87.21 & 89.41 & 91.05 & 92.43 & \textbf{93.17} & 87.80 & 86.85 & 93.58 & 94.37 & 93.46 & 92.11 \\
        Textual Entailment & 2400 & 82.83 & \textbf{87.91} & 82.52 & 83.96 & 86.61 & 88.43 & 86.93 & \textbf{97.07} & 84.76 & 84.41 & 97.00 & 99.04 & 89.44 & 88.15 \\
        Title Generation & 1784 & 83.97 & 84.81 & 75.57 & 77.12 & \textbf{86.61} & 86.67 & 84.88 & \textbf{89.43} & 85.44 & 84.45 & 88.35 & 90.01 & 88.95 & 87.23 \\
        Word Analogy & 800 & \textbf{83.93} & 82.67 & 83.35 & 83.22 & 81.15 & 88.35 & 83.63 & \textbf{89.00} & 85.17 & 81.53 & 93.37 & 97.37 & 88.44 & 89.37 \\
        \bottomrule
    \end{tabularx}
    \caption{Mean BERTScore recall values for all task types for all models we experimented with (Column names are abbreviated and abbreviations can be found in Table~\ref{tab:abbreviation_table_models}, \textbf{BOLD} values represent maximum value in pre-trained and instruction-tuned models groups respectively).}
    \label{tab:all_task_type_model_results}
\end{sidewaystable}

\begin{sidewaystable}
    \small
    \centering
    \begin{tabularx}{\textwidth}{
        >{\hsize=1.2\hsize}X | 
        >{\hsize=0.2\hsize}X | 
        >{\hsize=0.2\hsize}X  
        >{\hsize=0.2\hsize}X  
        >{\hsize=0.2\hsize}X  
        >{\hsize=0.2\hsize}X  
        >{\hsize=0.2\hsize}X | 
        >{\hsize=0.2\hsize}X  
        >{\hsize=0.2\hsize}X  
        >{\hsize=0.2\hsize}X  
        >{\hsize=0.2\hsize}X  
        >{\hsize=0.2\hsize}X | 
        >{\hsize=0.2\hsize}X  
        >{\hsize=0.2\hsize}X  
        >{\hsize=0.2\hsize}X  
        >{\hsize=0.2\hsize}X  
    }
        \toprule
        \textbf{Task Type} & \textbf{\# Inst.} & \multicolumn{5}{c}{\textbf{Pre-trained Models}} & \multicolumn{5}{c}{\textbf{Instruction-tuned Models}} & \multicolumn{4}{c}{\textbf{SOTA Models}} \\
        \cmidrule(lr){3-7} \cmidrule(lr){8-12} \cmidrule(lr){13-16}
        & & \textbf{S-1.7B} & \textbf{G-2B} & \textbf{G-7B} & \textbf{L-3-8B} & \textbf{F-2-11B} & \textbf{S-1.7B-I} & \textbf{G-2B-I} & \textbf{M-7B-I} & \textbf{G-7B-I} & \textbf{L-3-8B-I} & \textbf{GPT-4o-m} & \textbf{GPT-4o} & \textbf{DS-2} & \textbf{Gem-1.5-Pro} \\
        \midrule
        Abductive & 200 & 85.91 & 87.58 & 86.86 & 86.08 & \textbf{89.67} & 90.51 & 88.40 & \textbf{91.36} & 87.54 & 87.43 & 91.12 & 92.11 & 91.08 & 90.71 \\
        Analogical & 900 & \textbf{83.92} & 83.02 & 83.71 & 83.76 & 81.73 & 84.05 & 88.65 & \textbf{89.15} & 85.32 & 81.92 & 93.07 & 96.95 & 88.91 & 89.58 \\
        Causal & 800 & 83.62 & \textbf{87.00} & 81.19 & 79.85 & 86.90 & 85.21 & 85.18 & \textbf{93.35} & 84.78 & 84.65 & 95.58 & 95.88 & 90.16 & 89.35 \\
        Commonsense & 3000 & 83.80 & \textbf{86.91} & 83.75 & 82.26 & 85.61 & 87.31 & 88.96 & \textbf{94.31} & 85.07 & 83.89 & 95.65 & 97.82 & 91.35 & 89.60 \\
        Cross-document & 200 & 83.82 & \textbf{94.98} & 88.45 & 80.34 & 92.52 & 92.88 & 92.94 & \textbf{95.33} & 84.15 & 84.73 & 96.24 & 95.53 & 89.10 & 88.06 \\
        Deductive & 200 & 82.97 & \textbf{91.14} & 78.63 & 85.61 & 88.22 & 84.58 & 89.78 & \textbf{92.70} & 83.73 & 84.49 & 91.94 & 94.75 & 88.18 & 88.60 \\
        Discrete & 200 & 83.82 & \textbf{94.98} & 88.45 & 80.34 & 92.52 & 92.88 & 92.94 & \textbf{95.33} & 84.15 & 84.73 & 96.23 & 95.53 & 89.10 & 88.06 \\
        Logical & 100 & 80.87 & \textbf{83.58} & 77.46 & 80.16 & 79.78 & 82.16 & 86.13 & \textbf{96.19} & 81.19 & 80.50 & 97.26 & 99.12 & 88.09 & 87.42 \\
        Multihop & 226 & 81.65 & \textbf{94.98} & 86.04 & 79.84 & 92.07 & 88.62 & 92.94 & \textbf{94.11} & 83.22 & 83.82 & 92.52 & 94.08 & 89.01 & 86.26 \\
        Numerical & 200 & 83.82 & \textbf{94.98} & 88.45 & 80.34 & 92.52 & 92.88 & 92.94 & \textbf{95.33} & 84.15 & 84.73 & 96.24 & 95.53 & 89.10 & 88.06 \\
        Quantitative & 300 & 83.38 & \textbf{88.67} & 83.60 & 78.10 & 86.59 & 84.11 & 84.59 & \textbf{93.96} & 84.80 & 84.46 & 94.26 & 95.19 & 89.15 & 87.64 \\
        Reasoning on Actions & 300 & 84.96 & 85.03 & 82.07 & 86.80 & \textbf{89.73} & 85.94 & 87.18 & \textbf{92.31} & 86.03 & 85.12 & 89.95 & 92.78 & 92.48 & 89.37 \\
        Reasoning on Objects & 100 & 88.44 & 90.79 & 86.00 & 91.66 & \textbf{92.66} & 92.44 & 91.81 & \textbf{93.08} & 87.16 & 87.68 & 93.95 & 93.64 & 93.12 & 93.25 \\
        Social Interactions & 700 & 82.44 & 84.63 & 83.36 & \textbf{86.33} & 86.14 & 86.44 & 87.61 & \textbf{93.91} & 84.52 & 84.33 & 95.57 & 96.55 & 92.32 & 91.24 \\
        Relational & 1100 & 83.79 & 83.31 & \textbf{83.53} & 84.22 & 82.43 & 83.94 & 87.98 & \textbf{88.80} & 85.08 & 82.36 & 92.40 & 95.56 & 88.97 & 89.57 \\
        Temporal & 300 & 86.34 & \textbf{94.32} & 89.83 & 84.07 & 92.07 & 93.98 & 93.10 & \textbf{96.40} & 86.78 & 87.61 & 96.36 & 96.91 & 91.22 & 90.54 \\
        Textual Entailment & 2400 & 83.35 & \textbf{88.42} & 82.27 & 83.87 & 87.41 & 87.45 & 88.59 & \textbf{97.07} & 85.20 & 85.01 & 96.64 & 98.43 & 89.48 & 88.71 \\
        \bottomrule
    \end{tabularx}
    \caption{Mean BERTScore recall values for various reasoning types across different models (Column names are abbreviated and abbreviations can be found in Table~\ref{tab:abbreviation_table_models}, \textbf{BOLD} values represent the maximum value in pre-trained and instruction-tuned models groups respectively).}
    \label{tab:all_reasoning_model_results}
\end{sidewaystable}

\clearpage

\begin{sidewaystable}
    \small
    \centering
    \begin{tabularx}{\textwidth}{
        >{\hsize=1.2\hsize}X | 
        >{\hsize=0.2\hsize}X | 
        >{\hsize=0.2\hsize}X  
        >{\hsize=0.2\hsize}X  
        >{\hsize=0.2\hsize}X  
        >{\hsize=0.2\hsize}X  
        >{\hsize=0.2\hsize}X | 
        >{\hsize=0.2\hsize}X  
        >{\hsize=0.2\hsize}X  
        >{\hsize=0.2\hsize}X  
        >{\hsize=0.2\hsize}X  
        >{\hsize=0.2\hsize}X | 
        >{\hsize=0.2\hsize}X  
        >{\hsize=0.2\hsize}X  
        >{\hsize=0.2\hsize}X  
        >{\hsize=0.2\hsize}X  
    }
        \toprule
        \textbf{Task Type} & \textbf{\# Inst.} & \multicolumn{5}{c}{\textbf{Pre-trained Models}} & \multicolumn{5}{c}{\textbf{Instruction-tuned Models}} & \multicolumn{4}{c}{\textbf{SOTA Models}} \\
        \cmidrule(lr){3-7} \cmidrule(lr){8-12} \cmidrule(lr){13-16}
        & & \textbf{S-1.7B} & \textbf{G-2B} & \textbf{G-7B} & \textbf{L-3-8B} & \textbf{F-2-11B} & \textbf{S-1.7B-I} & \textbf{G-2B-I} & \textbf{M-7B-I} & \textbf{G-7B-I} & \textbf{L-3-8B-I} & \textbf{GPT-4o-m} & \textbf{GPT-4o} & \textbf{DS-2} & \textbf{Gem-1.5-Pro} \\
        \midrule
        Anthropology & 200 & 91.68 & 92.87 & 91.88 & \textbf{93.18} & 92.51 & 95.67 & 95.36 & \textbf{97.22} & 92.75 & 93.21 & 96.62 & 99.21 & 97.23 & 95.73 \\
        Books & 300 & 84.92 & \textbf{92.14} & 86.18 & 89.65 & 87.28 & 91.42 & 94.19 & \textbf{96.14} & 86.23 & 85.14 & 96.30 & 96.33 & 91.09 & 89.12 \\
        Captions & 700 & 83.94 & \textbf{92.01} & 87.52 & 86.04 & 86.05 & 90.18 & 93.16 & \textbf{94.11} & 85.93 & 84.90 & 98.08 & 98.58 & 89.54 & 89.62 \\
        Code & 100 & 82.36 & 85.24 & 86.76 & 88.73 & 84.31 & 86.23 & 98.87 & \textbf{99.10} & 85.26 & 83.26 & 99.84 & 99.87 & 89.79 & 89.39 \\
        Commonsense & 2500 & 84.33 & \textbf{85.00} & 80.50 & 81.26 & 84.29 & 85.82 & 88.77 & \textbf{94.03} & 85.54 & 83.60 & 95.86 & 97.82 & 91.51 & 90.84 \\
        Computer Science & 100 & 81.80 & 84.76 & 82.50 & 80.65 & \textbf{84.83} & 84.21 & 87.58 & \textbf{88.92} & 85.56 & 84.23 & 86.99 & 89.30 & 89.12 & 85.16 \\
        Debatepedia & 100 & 84.44 & \textbf{91.13} & 76.56 & 87.98 & 87.75 & 83.90 & 86.02 & \textbf{99.18} & 84.16 & 85.73 & 92.95 & 99.99 & 87.49 & 86.06 \\
        Dialogue & 1900 & 82.97 & \textbf{89.15} & 84.72 & 86.33 & 88.17 & 88.49 & 89.21 & \textbf{96.53} & 84.47 & 84.24 & 96.33 & 98.00 & 90.56 & 89.25 \\
        Economics & 100 & 83.31 & 82.17 & 82.92 & 80.23 & \textbf{94.22} & 84.46 & 87.85 & \textbf{97.90} & 84.88 & 86.14 & 84.54 & 99.91 & 99.41 & 88.57 \\
        English Exams & 100 & 90.75 & \textbf{93.52} & 86.47 & 89.76 & 87.45 & 94.29 & 93.72 & \textbf{95.29} & 89.25 & 87.84 & 95.67 & 95.79 & 94.51 & 94.13 \\
        Fiction & 700 & 86.53 & \textbf{93.22} & 90.11 & 93.25 & \textbf{93.39} & 94.50 & 93.53 & \textbf{99.13} & 88.06 & 87.86 & 97.15 & 99.66 & 92.54 & 89.63 \\
        Food & 200 & 82.67 & 84.96 & \textbf{91.03} & 89.24 & 84.15 & 86.30 & 86.07 & \textbf{93.16} & 84.44 & 84.82 & 94.84 & 95.94 & 91.87 & 89.78 \\
        Formal logic & 100 & 83.43 & 82.49 & 83.44 & \textbf{89.21} & 83.58 & 83.34 & 93.98 & \textbf{99.84} & 84.36 & 84.64 & 99.89 & 99.86 & 85.37 & 89.79 \\
        Government and Politics & 800 & 84.78 & 90.65 & 86.48 & 91.09 & \textbf{92.27} & 90.46 & 90.52 & \textbf{94.87} & 86.56 & 85.86 & 93.74 & 96.06 & 90.30 & 88.34 \\
        History & 800 & 87.98 & \textbf{93.89} & 90.68 & 93.66 & \textbf{94.60} & 95.58 & 93.36 & \textbf{98.58} & 89.66 & 89.89 & 96.67 & 99.51 & 93.08 & 91.22 \\
        Justice & 200 & 91.68 & 92.87 & 91.88 & \textbf{93.18} & 92.51 & 95.67 & 95.36 & \textbf{97.22} & 92.75 & 93.21 & 96.63 & 99.21 & 97.23 & 95.73 \\
        Knowledge Base & 100 & 85.91 & \textbf{89.95} & 84.38 & 85.13 & 87.31 & 91.83 & 90.15 & \textbf{92.69} & 87.73 & 86.26 & 91.96 & 92.10 & 92.23 & 92.17 \\
    \end{tabularx}
\end{sidewaystable}

\begin{sidewaystable}
    \centering
    \small
    \begin{tabularx}{\textwidth}{
        >{\hsize=1.2\hsize}X | 
        >{\hsize=0.2\hsize}X | 
        >{\hsize=0.2\hsize}X  
        >{\hsize=0.2\hsize}X  
        >{\hsize=0.2\hsize}X  
        >{\hsize=0.2\hsize}X  
        >{\hsize=0.2\hsize}X | 
        >{\hsize=0.2\hsize}X  
        >{\hsize=0.2\hsize}X  
        >{\hsize=0.2\hsize}X  
        >{\hsize=0.2\hsize}X  
        >{\hsize=0.2\hsize}X | 
        >{\hsize=0.2\hsize}X  
        >{\hsize=0.2\hsize}X  
        >{\hsize=0.2\hsize}X  
         >{\hsize=0.2\hsize}X  
    }
        Law & 700 & 86.59 & 94.23 & 89.15 & 92.57 & \textbf{94.85} & 93.79 & 93.24 & \textbf{97.77} & 88.26 & 95.29 & 96.52 & 98.20 & 91.49 \\
        Linguistics & 100 & 82.36 & 85.24 & 86.76 & 88.73 & 84.31 & 86.23 & 98.87 & \textbf{99.10} & 85.26 & 83.26 & 99.84 & 99.87 & 89.79 & 89.39 \\
        Miscellaneous & 800 & 84.23 & \textbf{88.27} & 85.86 & 82.09 & 85.67 & 85.66 & 87.06 & \textbf{96.43} & 85.44 & 84.83 & 96.64 & 97.52 & 91.03 & 89.71 \\
        Movies & 100 & 84.71 & \textbf{97.80} & 94.69 & 91.53 & 91.76 & \textbf{98.77} & 95.14 & 95.86 & 83.95 & 87.34 & 99.97 & 99.97 & 95.89 & 90.49 \\
        Narrative & 800 & 82.90 & 85.40 & 81.93 & 75.63 & \textbf{85.82} & 85.98 & 87.21 & \textbf{91.35} & 83.81 & 84.12 & 91.21 & 93.74 & 89.33 & 88.12 \\
        Natural Science & 400 & 84.44 & 84.77 & 81.85 & 86.30 & \textbf{91.05} & 85.84 & 89.25 & \textbf{95.15} & 86.39 & 84.98 & 94.86 & 96.41 & 88.16 & 89.92 \\
        News & 726 & 85.69 & \textbf{88.13} & 79.73 & 86.21 & 87.47 & 86.99 & 89.14 & \textbf{90.97} & 87.35 & 86.59 & 89.80 & 91.36 & 90.89 & 89.06 \\
        Nutrition & 100 & 83.24 & 86.16 & \textbf{94.79} & 92.12 & 82.84 & 88.81 & 90.11 & \textbf{96.79} & 84.39 & 86.16 & 99.97 & 99.98 & 93.33 & 90.50 \\
        Professions & 100 & 83.19 & 92.14 & 87.32 & 83.65 & 82.81 & 87.89 & 94.68 & \textbf{94.87} & 85.49 & 82.86 & 91.16 & 99.25 & 96.74 & 90.35 \\
        Public Places & 300 & 86.32 & \textbf{88.21} & 83.44 & 87.58 & 87.28 & 88.08 & 88.96 & \textbf{90.07} & 86.43 & 85.63 & 91.35 & 94.92 & 90.18 & 90.94 \\
        Reviews & 300 & 85.04 & 86.18 & 79.74 & 85.11 & \textbf{87.55} & 87.39 & 87.51 & \textbf{89.07} & 86.13 & 85.34 & 88.19 & 89.59 & 89.51 & 87.78 \\
        School Science Textbooks & 200 & 91.68 & 92.87 & 91.88 & \textbf{93.18} & 92.51 & 95.67 & 95.36 & \textbf{97.22} & 92.75 & 93.21 & 96.63 & 99.21 & 97.23 & 95.73 \\
        Scientific Research Papers & 400 & 82.41 & 84.26 & 81.28 & 80.06 & \textbf{85.72} & 86.13 & \textbf{86.89} & 86.84 & 85.22 & 83.94 & 89.29 & 89.88 & 89.77 & 85.34 \\
        Social Media & 200 & 84.04 & 85.52 & 78.21 & 89.79 & \textbf{91.51} & 86.19 & 90.87 & \textbf{96.03} & 85.43 & 86.06 & 95.75 & 96.09 & 89.48 & 90.06 \\
        Sports & 26 & 79.10 & 78.82 & 80.50 & 82.61 & 82.65 & 77.75 & 80.01 & \textbf{89.41} & 81.89 & 84.15 & 85.08 & 88.52 & 88.67 & 82.65 \\
        Statistics & 26 & 79.10 & 78.82 & 80.50 & 82.61 & 82.65 & 77.75 & 80.01 & \textbf{89.41} & 81.89 & 84.15 & 85.08 & 88.52 & 88.67 & 82.65 \\
        Story & 500 & 83.73 & \textbf{89.49} & 82.21 & 67.79 & 89.23 & 84.78 & 88.29 & \textbf{92.66} & 85.25 & 85.18 & 90.67 & 93.81 & 89.98 & 88.25 \\
        Web & 400 & 86.67 & 87.18 & 83.53 & 82.49 & \textbf{90.95} & 89.79 & 90.17 & \textbf{94.90} & 86.65 & 86.78 & 95.22 & 97.31 & 91.62 & 90.46 \\
        Wikipedia & 2184 & 84.60 & 86.74 & 80.90 & 84.41 & \textbf{88.66} & 88.75 & 89.03 & \textbf{95.12} & 85.72 & 85.78 & 94.46 & 96.03 & 92.27 & 90.45 \\
        \bottomrule
    \end{tabularx}
    \caption{Mean BERTScore recall values for various application domains across different models (Column names are abbreviated and abbreviations can be found in Table~\ref{tab:abbreviation_table_models}, \textbf{BOLD} values represent maximum value in pre-trained and instruction-tuned models groups respectively).}
    \label{tab:all_domain_model_results}
\end{sidewaystable}

\clearpage
\section{LM-level Results}
\label{app:lm_level_results}

\begin{table*}[htbp]
\centering
\begin{tabular}{l|c|cc|ccccccc}
\toprule
\textbf{Model} & \textbf{\# Params} & \textbf{Def} & \textbf{Ex.} & \textbf{R-1} & \textbf{R-2} & \textbf{R-L} & \textbf{METEOR} & \textbf{BERTScore P/R/F1} \\ 
\toprule
SmolLM-1.7B & 1.71B & \checkmark & 2 & 2.50 & 1.07 & 2.27 & 4.92 & 67.34/83.71/74.56 \\
Gemma-2B & 2.51B & \checkmark & 4 & \textbf{22.04} & \textbf{7.88} & \textbf{21.23} & \textbf{18.12} & 78.22/\textbf{86.41}/81.88 \\
Gemma-2-2B & 2.61B & \checkmark & 0 & 7.56 & 2.18 & 7.21 & 9.43 & 70.29/83.66/76.23 \\
Mistral-7B & 7.25B & \checkmark & 8 & 1.17 & 0.54 & 1.08 & 1.99 & 49.25/58.41/53.40 \\
Gemma-7B & 8.54B & \checkmark & 0 & 18.17 & 5.89 & 17.49 & 16.14 & 71.86/81.06/75.94 \\
Llama-3-8B & 8.03B & \checkmark & 0 & 16.38 & 5.35 & 15.30 & 14.96 & 75.52/82.73/78.80 \\
Falcon-2-11B & 11.1B & \checkmark & 8 & 16.88 & 6.46 & 16.01 & 16.45 & 79.65/86.18/82.72 \\ 
\midrule
SmolLM-1.7B-I & 1.71B & \checkmark & 2 & 20.22 & 7.59 & 19.03 & 18.78 &80.34/86.66/83.24 \\
Gemma-2B-I & 2.51B & \checkmark & 2 & 27.56 & 8.08 & 26.24 & 20.62 & 84.56/88.06/86.19 \\
Gemma-2-2B-I & 2.61B & \texttimes & 0 & 3.45 & 1.35 & 2.99 & 5.60 & 73.91/82.76/78.05 \\
Mistral-7B-I & 7.25B & \checkmark & 8 & \textbf{51.96} & \textbf{14.67} & \textbf{50.12} & \textbf{35.55} & 91.29/\textbf{93.76}/92.39 \\
Gemma-7B-I & 8.54B & \checkmark & 0 & 8.64 & 3.23 & 7.96 & 12.57 & 78.18/85.14/81.48 \\
Llama-3-8B-I & 8.03B & \texttimes & 8 & 4.68 & 2.19 & 4.23 & 8.31 & 74.23/84.33/78.89 \\
\bottomrule
\end{tabular}
\caption{Mean Performance Metrics of Models with \textbf{\# Params} number of parameters. \textbf{Def} indicates presence (\checkmark) or absence (\texttimes) of task definitions, and \textbf{Ex.} shows the number of examples in the best overall prompt style. BERTScore P/R/F1 represents Precision, Recall, and F1 (\textbf{BOLD} represents best performing metrics in both model categories).}
\label{tab:all_metrics_results}
\end{table*}

In Table~\ref{tab:all_metrics_results}, we report the best prompt style at the LM-level, abstracting all analyses at aspect-level with different performance metrics like ROUGE 1/2/L~\citep{lin2004rouge}, METEOR~\citep{banerjee2005meteor} and BERTScore P/R/F1~\citep{zhang2019bertscore} for reference. 

Here we also include the results on Gemma-2-2B, Gemma-2-2B-I~\citep{gemma2} and Mistral-7B-v0.3~\citep{mistral}. From the results, it is visible why we ignored these models from main analysis. The Gemma-2 family is performing less compared to Gemma, and we wanted to keep a wide LM family for analysis. For Mistral, it was just underperforming. So, for brevity, we excluded them.

Among pre-trained models, Gemma-2B, the smallest of all models, gives best results. In IT models, Mistral-7B-I significantly outperforms others, despite its pre-trained version under-performing. This can be due of extensive fine-tuning of Mistral using several conversational datasets. 
\section{Prompt Analyses}

This appendix aims to analyze the performance of LMs on various prompts, offering an extension of the ideas discussed in the main paper.

\newcommand{\promptWisePerformanceLineGraph}[1]{
\begin{figure*}[htbp]
    \centering
    \begin{subfigure}[b]{0.325\linewidth}
        \includegraphics[width=\linewidth]{assets/results/experiment_4/#1/categories/categories_with_definition.pdf}
        \caption{Task Types; with definition}
    \end{subfigure}
    \hfill
    \begin{subfigure}[b]{0.325\linewidth}
        \includegraphics[width=\linewidth]{assets/results/experiment_4/#1/domains/domains_with_definition.pdf}
        \caption{Domains; with definition}
    \end{subfigure}
    \hfill
    \begin{subfigure}[b]{0.325\linewidth}
        \includegraphics[width=\linewidth]{assets/results/experiment_4/#1/reasoning/reasoning_with_definition.pdf}
        \caption{Reasonings; with definition}
    \end{subfigure}
    \newline
    \begin{subfigure}[b]{0.325\linewidth}
        \includegraphics[width=\linewidth]{assets/results/experiment_4/#1/categories/categories_without_definition.pdf}
        \caption{Task Types; without definition}
    \end{subfigure}
    \hfill
    \begin{subfigure}[b]{0.325\linewidth}
        \includegraphics[width=\linewidth]{assets/results/experiment_4/#1/domains/domains_without_definition.pdf}
        \caption{Domains; without definition}
    \end{subfigure}
    \hfill
    \begin{subfigure}[b]{0.325\linewidth}
        \includegraphics[width=\linewidth]{assets/results/experiment_4/#1/reasoning/reasoning_without_definition.pdf}
        \caption{Reasonings; without definition}
    \end{subfigure}
    \caption{Mean BERTScore recall variation for \textbf{#1} across various task types, domains, and reasoning types by varying number of in-context examples, segmented by with and without using task definitions (Columns: different aspects, row 1: with task definition, row 2: without task definitions).}
\end{figure*}
}

\subsection{Best Prompt Style at LM Level}
\label{app:best_prompt_style_lm_level}

We report BERTScore recall values for all prompt styles used in this work at Language Model level without going into the aspects in Table~\ref{tab:lm_level_performance_instruction}. These are scores on the entire experimental dataset.

\begin{table*}[htbp]
    \centering
    \begin{tabular}[width=\textwidth]{c|cccc|cccc}
         \toprule
        {\textbf{Model Name}} & \multicolumn{4}{c}{\textbf{With Definition}} & \multicolumn{4}{c}{\textbf{Without Definition}} \\
        \cmidrule(lr){2-5} \cmidrule(lr){6-9}
        & {0} & {2} & {4} & {8} & {0} & {2} & {4} & {8} \\
        \midrule
        SmolLM-1.7B         & 83.33  & 83.71  & 83.66  & 83.69  & 82.68  & 83.30  & 83.28  & 83.30  \\
        Gemma-2B          & 84.69 & 86.15 & 86.41 & 86.34 & 82.13 & 81.79 & 81.14 & 81.17 \\
        Gemma-7B          & 81.06 & 68.29 & 67.87 & 68.10 & 65.72 & 72.90 & 71.67 & 71.48 \\
        Meta-Llama-3-8B   & 82.73 & 52.43 & 52.13 & 52.45 & 77.98 & 56.17 & 54.18 & 53.30 \\
        Falcon-2-11B      & 84.27 & 86.06 & 86.05 & 86.18 & 83.46 & 85.61 & 86.06 & 86.09 \\
        SmolLM-1.7B-I       & 84.61  & 86.66  & 86.55  & 86.44  & 83.34  & 86.45  & 85.80  & 85.84  \\ 
        Gemma-2B-I        & 87.79 & 88.06 & 87.96 & 88.05 & 84.70 & 86.03 & 86.24 & 86.28 \\
        Mistral-7B-I & 88.29 & 93.04 & 93.75 & 93.76 & 83.82 & 88.88 & 90.20 & 90.28 \\
        Gemma-7B-I        & 85.14 & 84.71 & 84.76 & 84.82 & 83.58 & 83.96 & 84.08 & 84.05 \\
        Meta-Llama-3-8B-I & 84.11 & 84.11 & 84.04 & 83.96 & 82.79 & 84.30 & 84.25 & 84.33 \\
        \bottomrule
    \end{tabular}
    \caption{Mean BERTScore recall values of all LMs with different prompt styles on the entire experimental dataset (2nd-level column denotes number of examples of the prompt style).}
    \label{tab:lm_level_performance_instruction}
\end{table*}

From the table, we see that the differences with increasing examples are less prominent as compared to the aspect-level analyses of prompt style in Section~\ref{sec:comparison_prompts} and Appendix~\ref{app:prompt_line_graphs}. This highlights the importance of conducting the prompt style analysis at aspect level. It is important to determine the prompt style that serves the best for a given use-case.

\subsection{Variation of Performance with Different Prompt Styles for all Language Models}
\label{app:prompt_line_graphs}

This is a continuation from Section~\ref{sec:comparison_prompts} where we analyzed how performance of Mistral-7B-I varied for different task types, domains and reasoning types with the 8 different prompt styles that we use. In this section, we will provide similar visualizations for all other models. Using these graphs, one can determine the best prompt style for that particular task type, domain, or reasoning type. Additionally, the performance trade-off of using any other prompt style can also be analyzed. The visualizations are provided in Figures 8 -- 17. From these, it is clear that for each LM, the variation in performance is different for each entity of task type, application domain and reasoning type. Therefore, the prompt style should be carefully selected by examining the trend.

\subsection{Paraphrasing Definitions}
\label{app:paraphrasing_definitions}

In Section~\ref{sec:prompts} and Section~\ref{sec:comparison_paraphrased_def}, we discussed about paraphrasing the task definitions. Here, we give more details around how we did the paraphrasing. We also reported results for only four LMs in the main paper, but here, we will provide the performance change for all LMs. We use the following prompt to paraphrase task definitions with GPT-3.5-Turbo~\citep{gpt3,gpt35turbo} to generate paraphrases. Some paraphrases generated are given in Table~\ref{tab:paraphrased_def_examples}.

\begin{mdframed}
\texttt{You are an AI assistant designed to paraphrase a definition of a task. You will be provided with a paragraph that defines a particular task to be done. Your task is to paraphrase the given definition so that it is interpretable by another AI assistant to fulfill the task. Make sure to not omit any information from the paragraph. It might be necessary to complete the task. Only paraphrase it. \\
\{task\_definition\}
}  
\end{mdframed}

\begin{table*}[htbp]
\centering
    \begin{tabular}[width=\linewidth]{c|c|cc|c}
        \toprule
        \textbf{Model Name} & \textbf{Ex.} & \textbf{Def.} & \textbf{Par. Def.} & \textbf{\% Dec} \\
        \midrule
        SmolLM-1.7B & 2 & 83.71 & 83.17 & 0.54 \\
        Gemma-2B & 4 & 86.410 & 85.771 & 0.74 \\
        Gemma-7B & 0 & 81.055 & 80.998 & 0.07 \\
        Meta-Llama-3-8B & 0 & 82.727 & 82.501 & 0.27 \\
        Falcon-2-11B & 8 & 86.184 & 86.000 & \textbf{0.21} \\
        \midrule
        SmolLM-1.7B-I & 0 & 86.66 & 86.26 & 0.46 \\
        Gemma-2B-I & 4 & 87.959 & 87.671 & 0.33 \\
        Mistral-7B-I-v0.3 & 8 & 93.755 & 93.219 & 0.57 \\
        Gemma-7B-I & 0 & 85.142 & 84.825 & 0.37 \\
        Meta-Llama-3-8B-I & 0 & 84.112 & 84.217 & \textbf{0.12} \\
        \bottomrule
    \end{tabular}
\caption{Mean BERTScore recall values of outputs with actual task definition (Def) and paraphrased definitions (Par. Def), along with percentage decrease in value (\% Dec) when paraphrased definitions are used using `Ex.’ in-context examples for all models (\textbf{BOLD} values indicate least decrease in percentage in the two types of LMs).}
\label{tab:paraphrased_def_comparison_all_models}
\end{table*}

The mean BERTScore recall values of the performance of all the 10 models with actual and paraphrased definitions are given in Table~\ref{tab:paraphrased_def_comparison_all_models}. This will support the arguments in Section~\ref{sec:comparison_paraphrased_def}.

\begin{table*}[htbp]
\centering
    \begin{tabular}{>{\raggedright\arraybackslash}m{7.5cm}|>{\raggedright\arraybackslash}m{7.5cm}}
        \toprule
        \textbf{Given Task Definition} & \textbf{Paraphrased Task Definition} \\
        \hline
        The task is to write a full sentence or two using all of the information given. The sentence(s) will be a brief review of a restaurant. Use all of the information provided. & Create a concise statement that summarizes the overall experience and offering of a dining establishment, utilizing all the details provided. \\
        \midrule
        In this task you are given a story and a question regarding that story. You must judge whether the question is answerable based on the info given to you. Label the instances as "Answerable" or "Not Answerable" based on your judgment. the story and the question are separated by a new line character. & Your task is to determine whether a given question, presented alongside a corresponding story, is answerable. You will need to classify each instance as either "Answerable" or "Not Answerable" based on the information provided. The story and the question are presented separately, with a new line character separating them. \\
        \midrule
        In this task, you are given a context tweet, a question and corresponding answer of given question. Your task is to classify given passage into two categories: (1) "yes" if the given context is useful in answering the question, and (2) "no" if the given context is not useful. & Your job is to determine if a provided passage, including a context tweet, a question, and its answer, is helpful in answering the question by classifying it as either "yes" if the context is useful or "no" if it is not. \\
        \bottomrule
    \end{tabular}
\caption{Comparison between actual and paraphrased task definitions.}
\label{tab:paraphrased_def_examples}
\end{table*}

\subsection{Adversarial Definitions}
\label{app:adversarial_definitions}

This experiment aims to identify how robust the LMs are when they are asked to complete a task instance with a task definition that has subtle differences capable confuse it, or are provided to elicit a response that is not desired. These subtle differences can both be intentional and non-intentional.

To perform this, similar to Appendix~\ref{app:paraphrasing_definitions}, we generate adversarial task definitions for all the task definitions available in the dataset using \texttt{gpt-3.5-turbo}~\citep{gpt3,gpt35turbo} using a pre-determined prompt which instructs it to generate the adversarial definition. The prompt is given below, and some examples are given in Table~\ref{tab:adversarial_def_examples}.

\begin{mdframed}
\texttt{Consider the following task description: \{task\_definition\}. \\ Now, create an adversarial version of this task description. The adversarial task should appear legitimate but include subtle changes that could potentially mislead a language model or alter its output significantly. Focus on altering keywords, injecting ambiguities, or rephrasing the instructions in a way that shifts the expected response without being overtly misleading.
}  
\end{mdframed}

\begin{table*}[htbp]
\centering
    \begin{tabular}{>{\raggedright\arraybackslash}m{6cm}|>{\raggedright\arraybackslash}m{6cm}|>{\raggedright\arraybackslash}m{3cm}}
        \toprule
        \textbf{Given Task Definition} & \textbf{Adversarial Task Definition} & \textbf{Difference} \\
        \hline
        The task is to write a full sentence or two using all of the information given. The sentence(s) will be a brief review of a restaurant. Use all of the information provided. & Write a concise summary of a dining experience at a restaurant by incorporating all the details provided. Craft a couple of sentences that capture the essence of the establishment. Be sure to utilize all the data given effectively. & Dining experience is not always the complete review of the restaurant. \\
        \midrule
        In this task you are given a story and a question regarding that story. You must judge whether the question is answerable based on the info given to you. Label the instances as "Answerable" or "Not Answerable" based on your judgment. the story and the question are separated by a new line character. & In this task, you will be provided with a narrative followed by an inquiry. Your task is to determine the question's answerability based on the given information. You are required to classify the statements as either "Answerable" or "Not Answerable" based on your assessment. Please note that paragraphs containing the narrative and question shall be separated by a newline character. & Narrative and inquiry are not same as story and question. Additionally, it is not specified that inquiry is related to the narrative. \\
        \bottomrule
    \end{tabular}
\caption{Comparison between actual and adversarial task definitions.}
\label{tab:adversarial_def_examples}
\end{table*}

Then, we use the prompt style with definition and 0 examples, but replace the definition with the adversarial definition of the task. At last, we calculate the BERTScore recall values for adversarial versus actual task definition, and report the results in Table~\ref{tab:adversarial_def_comparison_all_models}. The reason to choose 0 examples was to avoid the scenario of the model recovering by learning from in-context examples.

\begin{table*}[htbp]
\centering
    \begin{tabular}{c|cc|c}
        \toprule
        \textbf{Model Name} & \textbf{Def.} &  \textbf{Adv. Def.} & \textbf{\% Dec.} \\
        \midrule
        SmolLM-1.7B & 83.33 & 82.21 & 1.34 \\
        Gemma-2B & 84.68 & 83.94 & 0.88 \\
        Gemma-7B & 81.06 & 78.67 & 2.94 \\
        Llama-3-8B & 82.73 & 78.01 & 5.70 \\
        Falcon-2-11B & 84.27 & 83.75 & \textbf{0.61} \\ \midrule
        SmolLM-1.7B-I & 84.61 & 83.38 & 1.46 \\
        Gemma-2B-I & 87.79 & 86.74 & 1.20 \\
        Mistral-7B-I & 88.29 & 86.90 & 1.58 \\
        Gemma-7B-I & 85.14 & 83.87 & 1.50 \\
        Llama-3-8B-I & 84.11 & 83.57 & \textbf{0.65} \\
        \bottomrule
    \end{tabular}
    \caption{Mean BERTScore recall values of outputs using actual task definition (Def.) versus adversarial definitions (Adv Def.) using 0 in-context examples for all models with percentage decrease (\% Dec.) in performance with adversarial definitions (\textbf{BOLD} values indicate least decrease in percentage in the two types of LMs).}
    \label{tab:adversarial_def_comparison_all_models}
\end{table*}

From the table, we see that most models are robust to adversarial change in task definition, with 7/10 models suffering less than 3\% of decrease in performance. For the pre-trained model, Falcon-2-11B is most robust with only 0.613\% decrease, and Gemma-2B ranks second with 0.88\% decrease. For IT models, Gemma-2B-I is still one of the best, suffering only 1.2\% decrease in BERTScore recall values only, but is outperformed by Llama-3-8B-I. Mistral-7B-I, the best performing IT model on true definitions is also not very sensitive to this change. Ph-3-mini-128k-I suffers the most amongst all models. We have seen sensitivity to be a general trend in this model with all varying parameters.

\promptWisePerformanceLineGraph{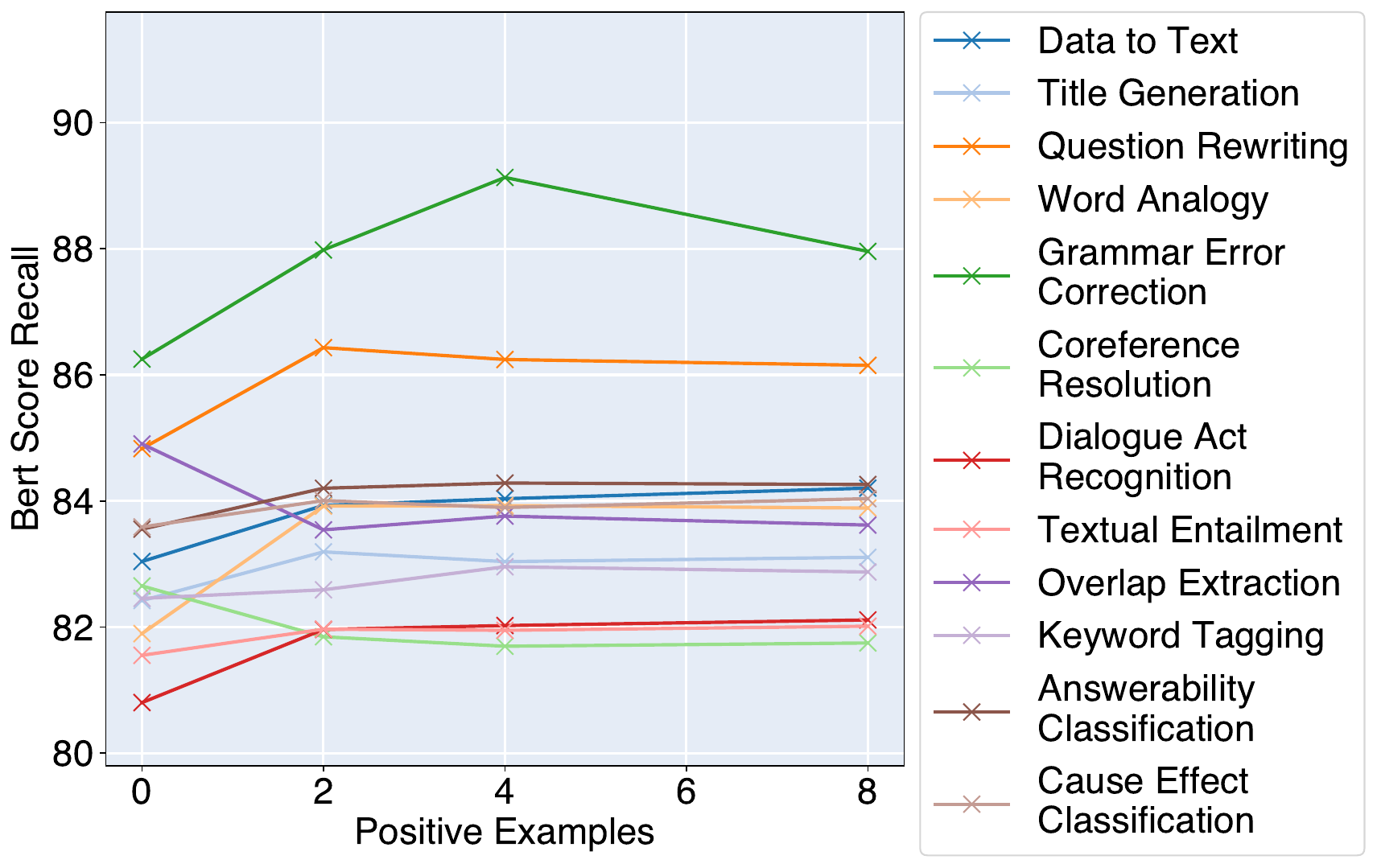}
\promptWisePerformanceLineGraph{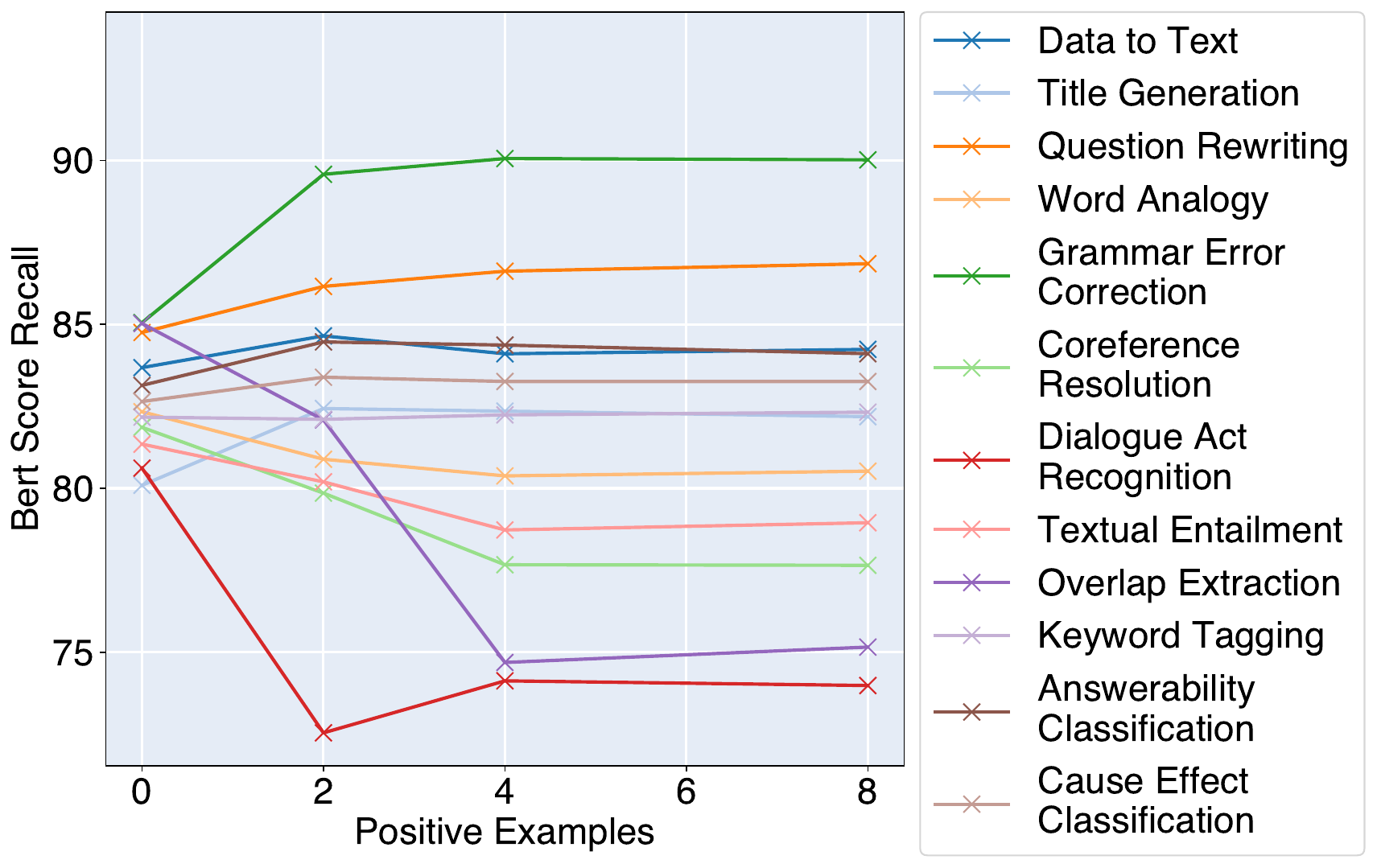}
\promptWisePerformanceLineGraph{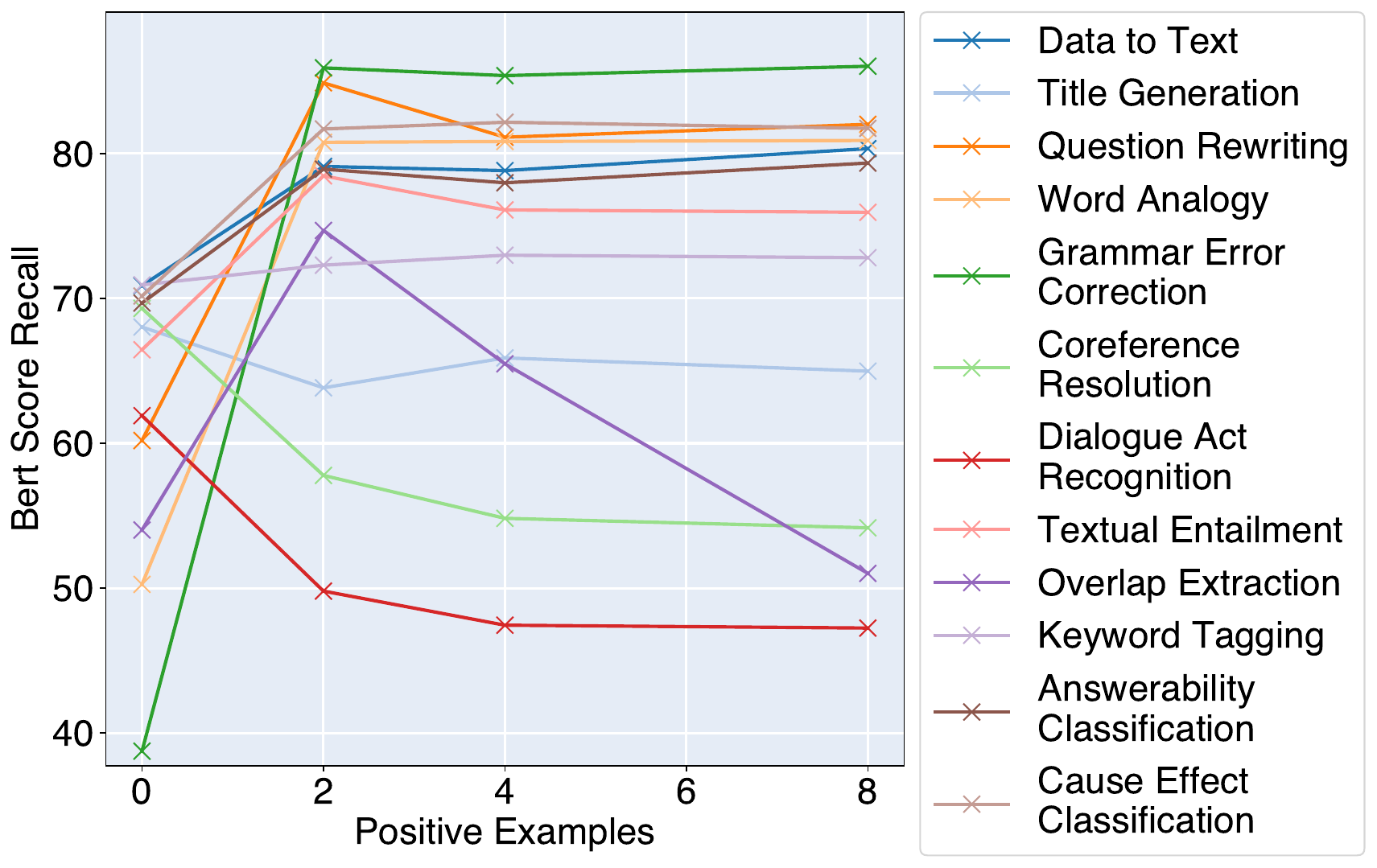}
\promptWisePerformanceLineGraph{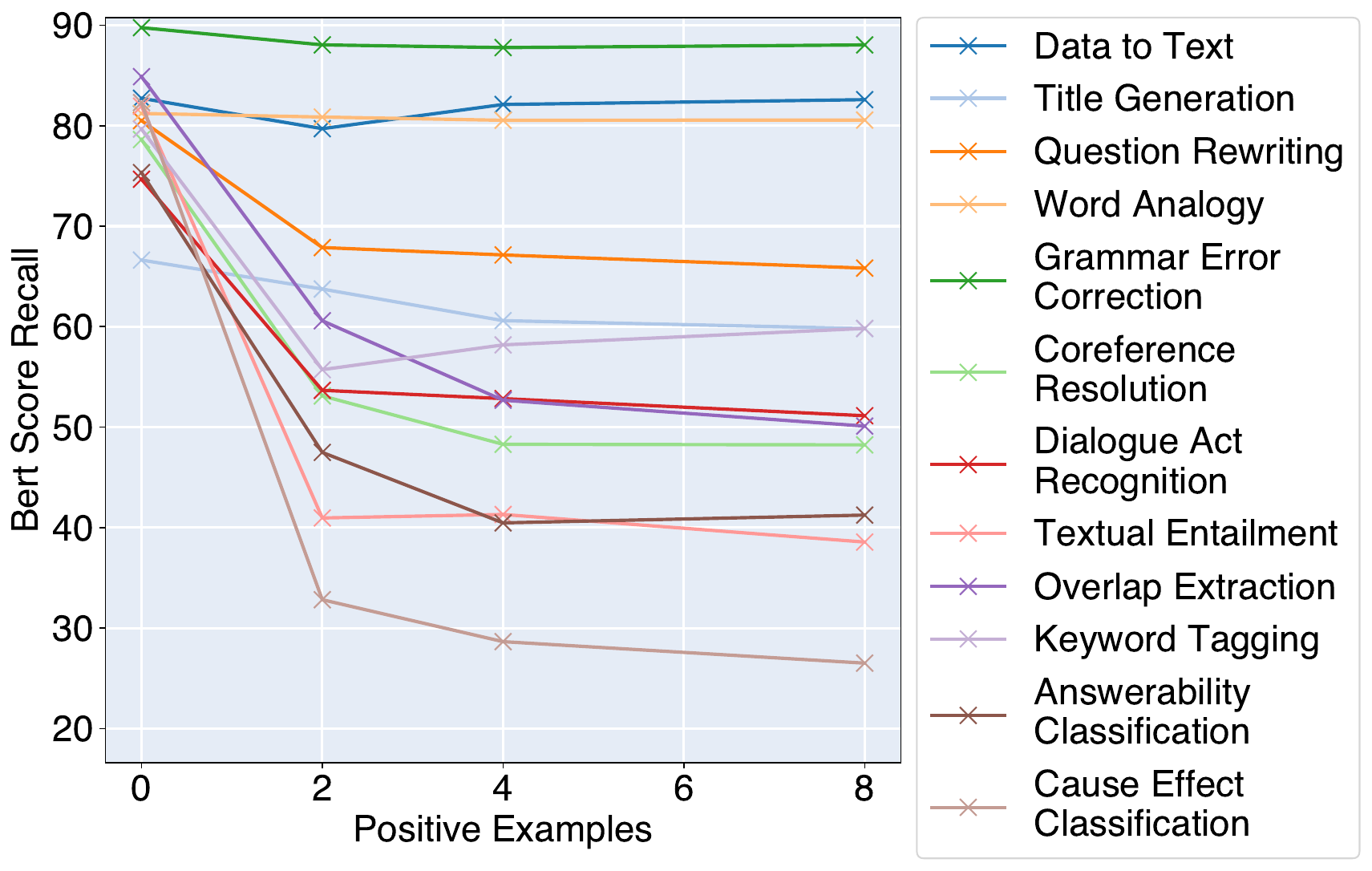}
\promptWisePerformanceLineGraph{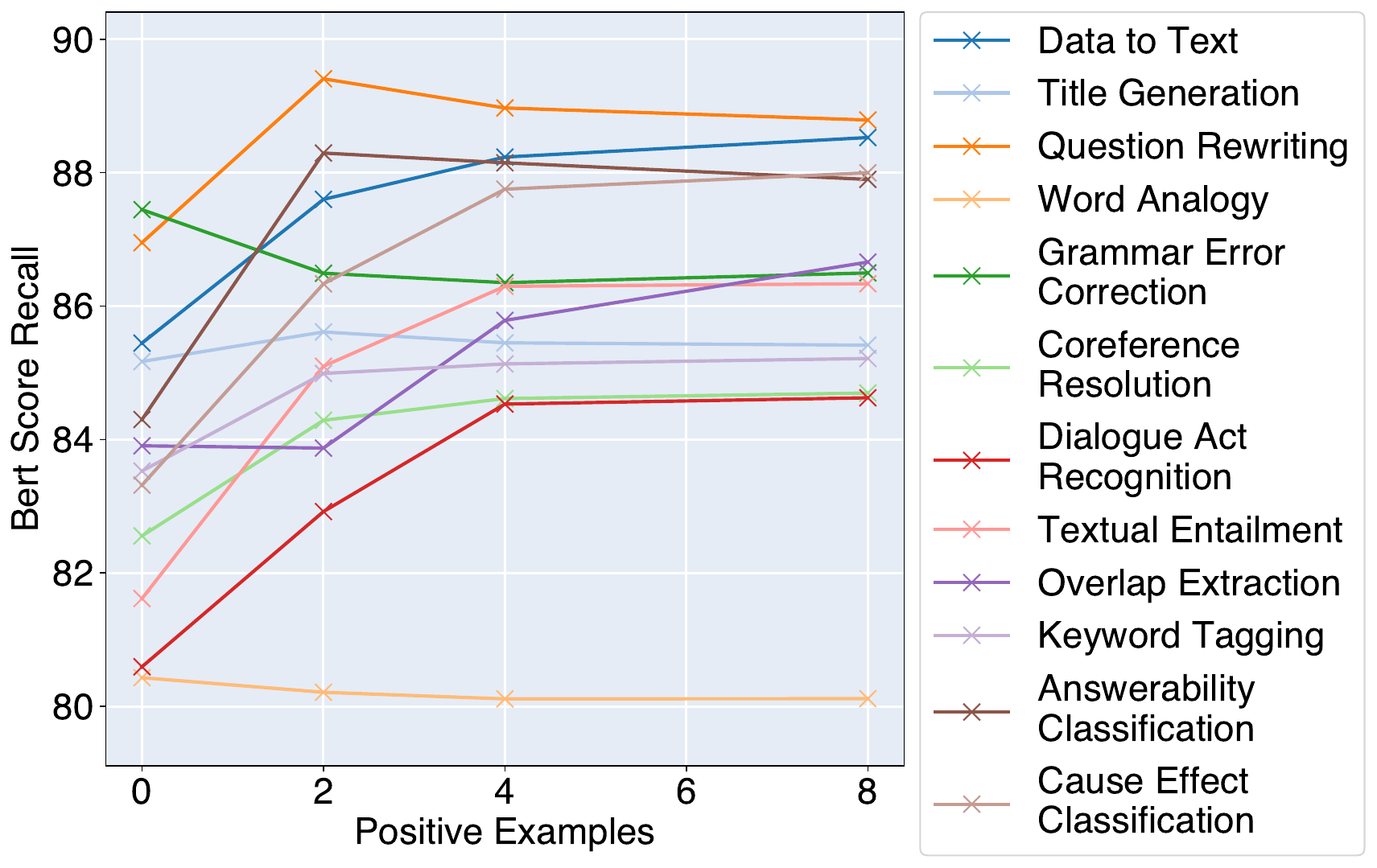}
\promptWisePerformanceLineGraph{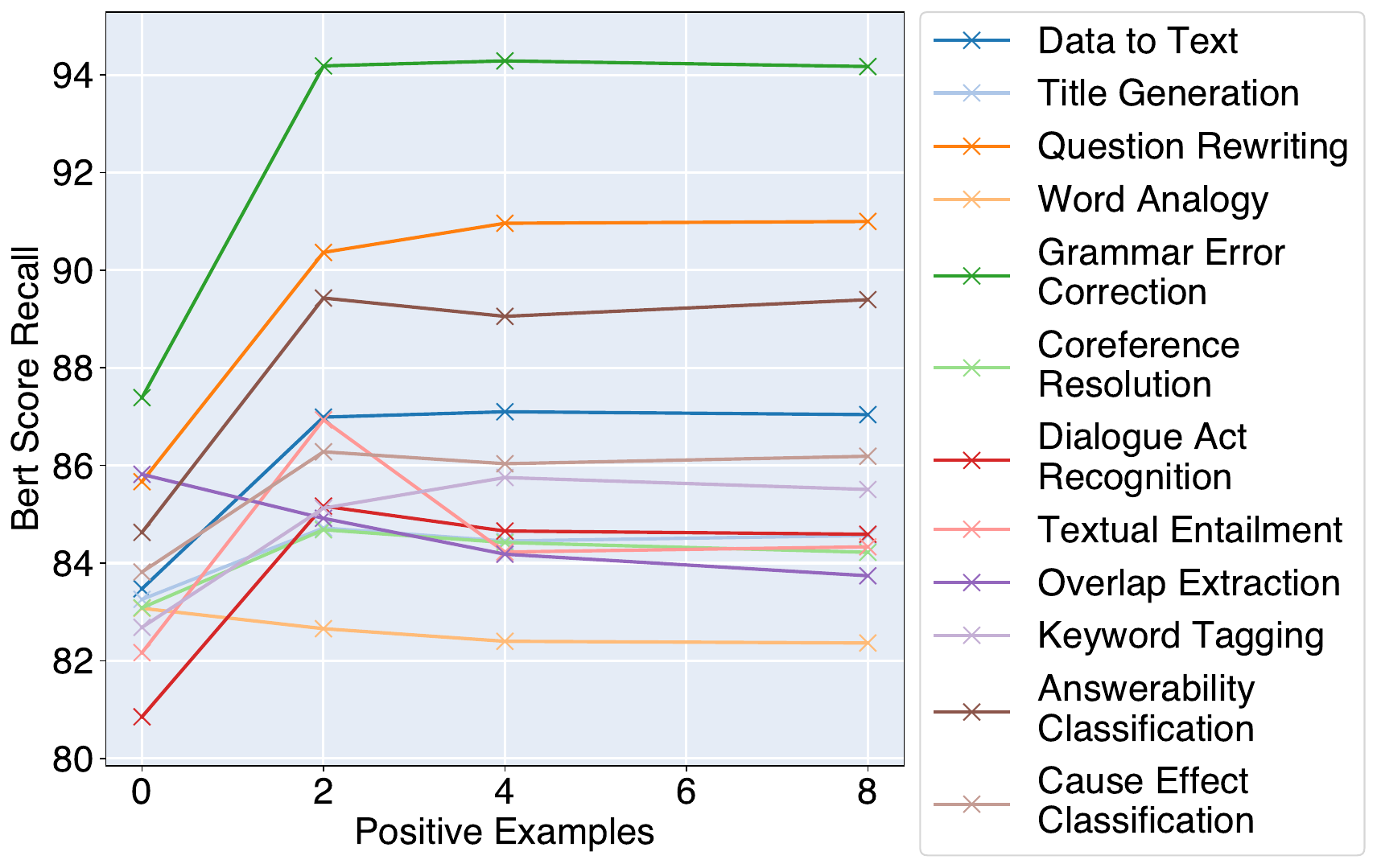}
\promptWisePerformanceLineGraph{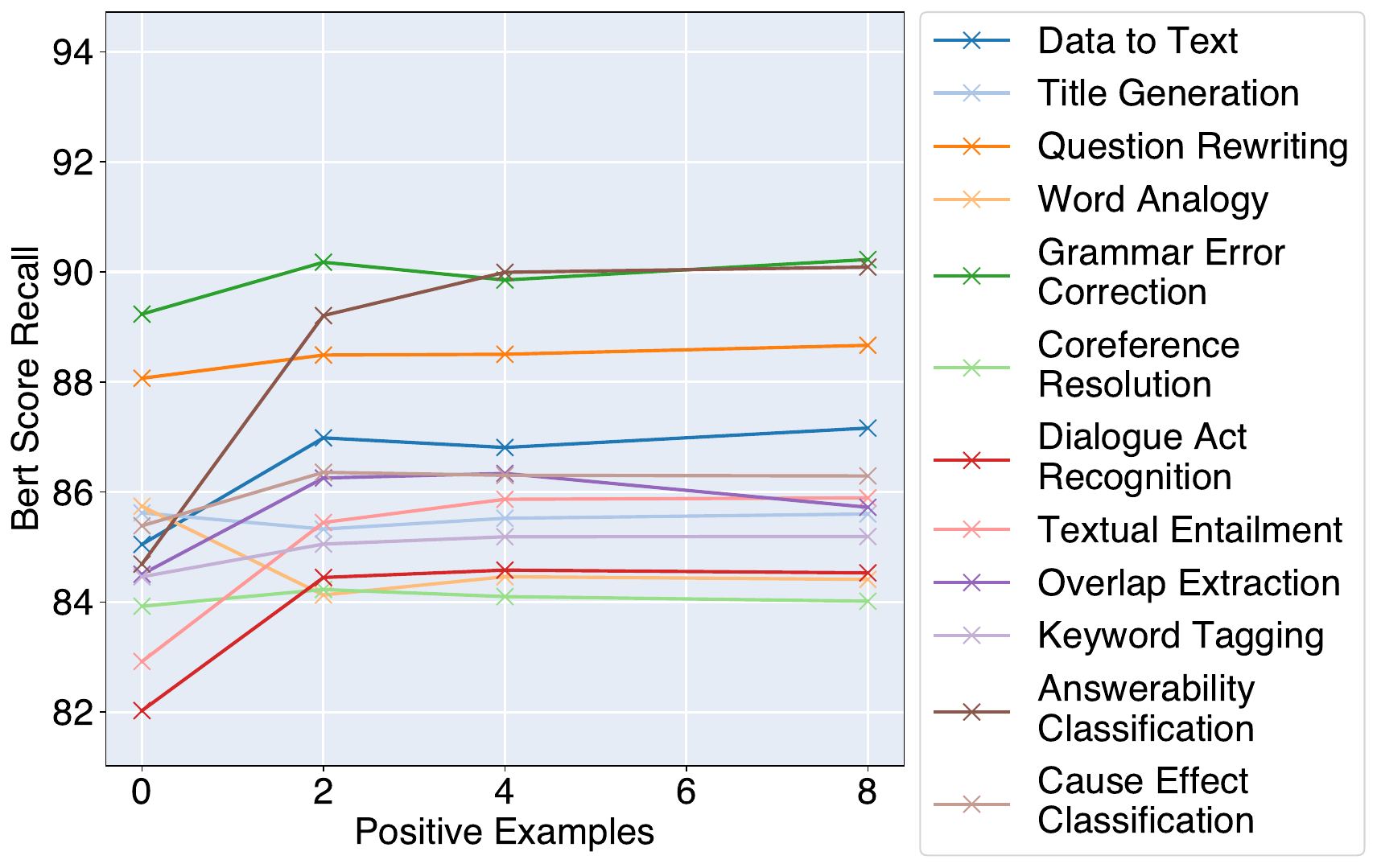}
\promptWisePerformanceLineGraph{Mistral-7B-I}
\promptWisePerformanceLineGraph{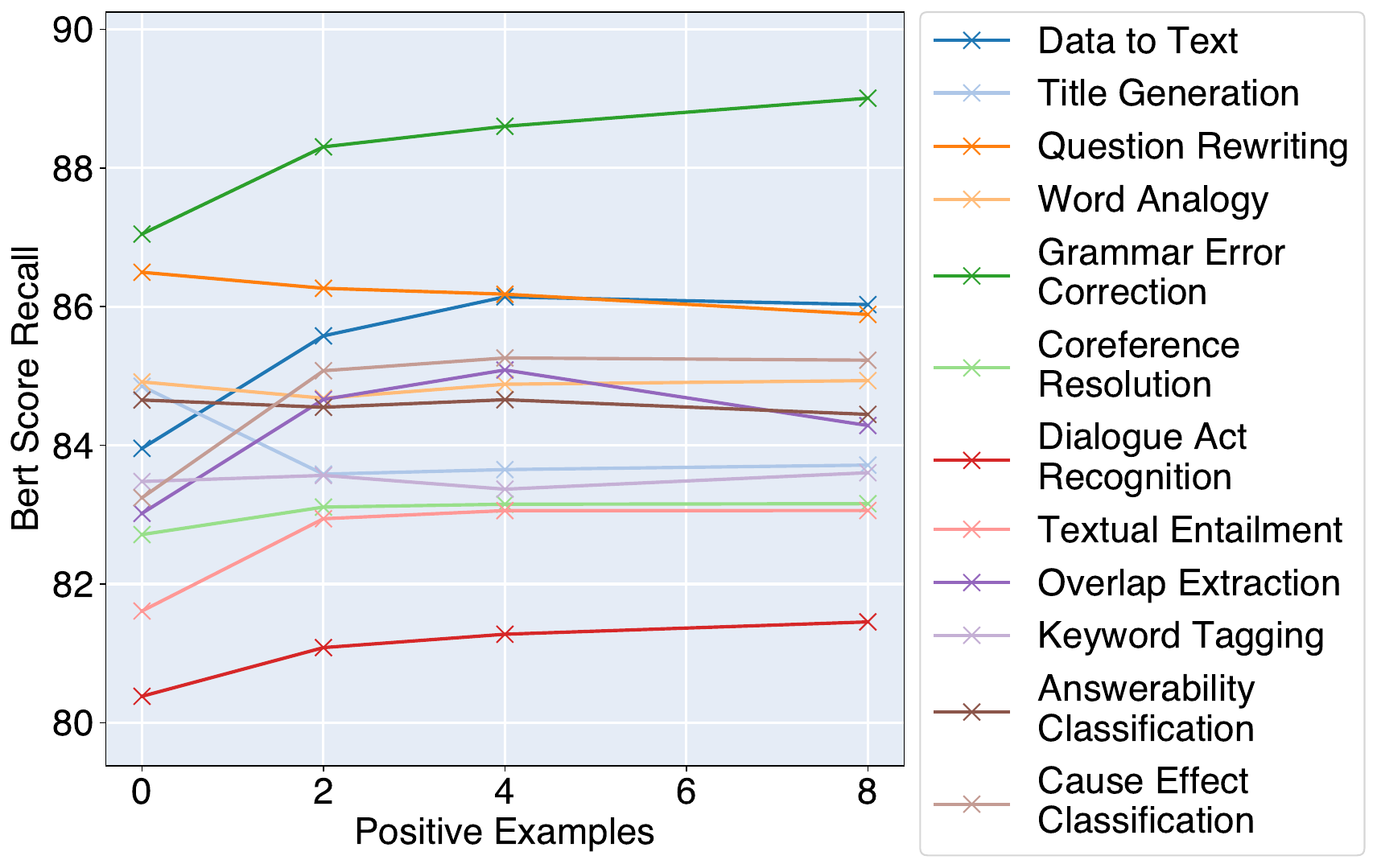}
\promptWisePerformanceLineGraph{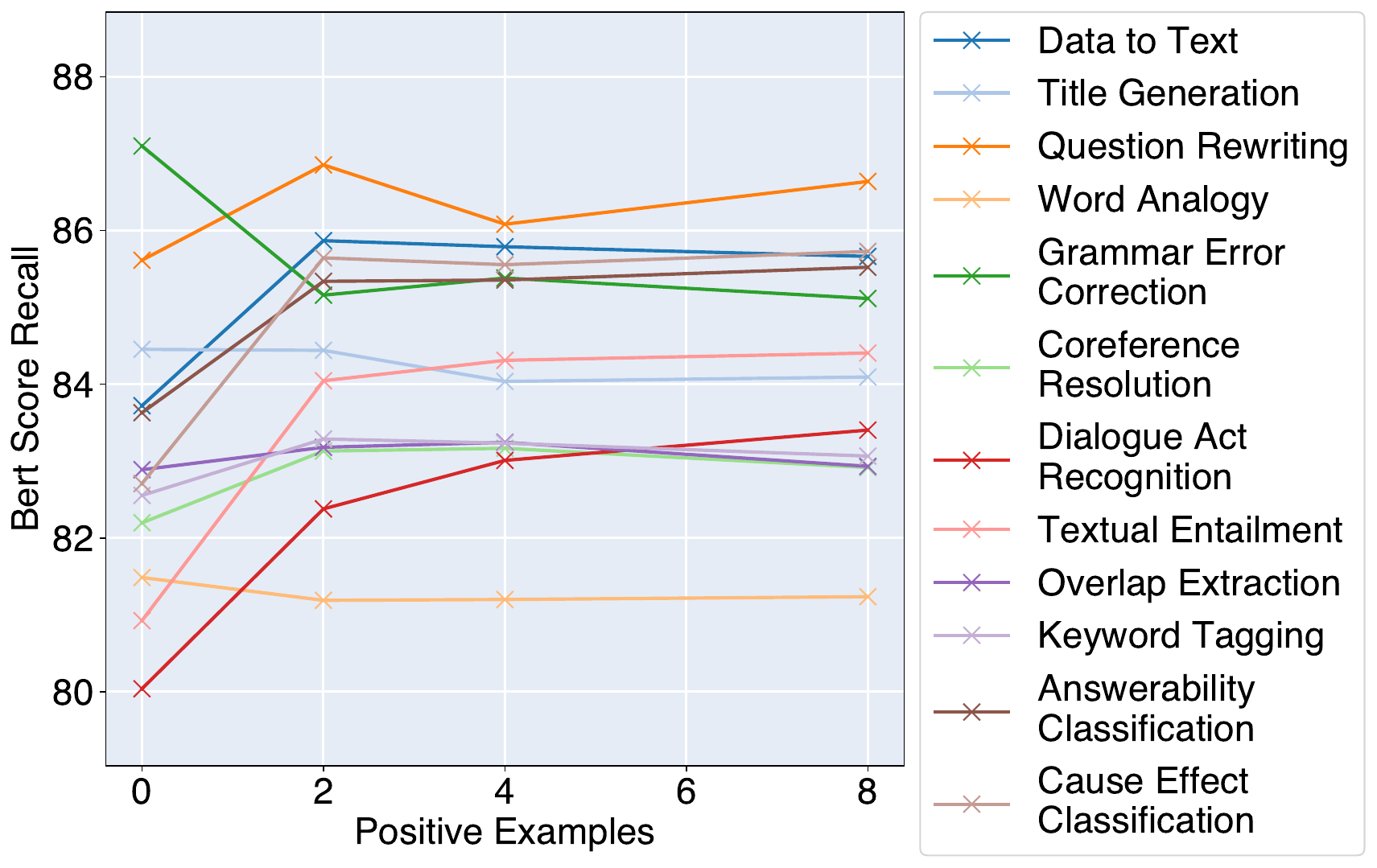}

\section{Comparison of Decoding Techniques}
\label{app:comparison_decoding}

In all experiments, we used greedy decoding. However, sometimes using top-k or top-p sampling~\citep{top_p_sampling} can offer better results. To analyze the impact of these sampling techniques, we generate and evaluate outputs with both these for each LM using the best instruction as per Table~\ref{tab:all_metrics_results}. The results are given in Table~\ref{tab:sampling_comparison}. We also tried different values of $k$ and $p$, but did not find any meaningful change in the results.

From the table, we can see that the performance doesn't change significantly at the LM level. In general, the performance of all LMs remains roundabout similar. We didn't observe a significant change in performance at aspect and entity level also. Given these factors, we preferred greedy decoding since it offers other advantages such as efficiency and reproducibility.

\begin{table*}[htbp]
    \centering
    \begin{tabular}{l|ccc}
    \toprule
        \textbf{Model Name} & \textbf{Greedy} & \textbf{top-k} & \textbf{top-p} \\
        & & $k=10$ & $p=0.9$ \\ \midrule
        SmolLM-1.7B & 83.71 & 83.72 & 83.71 \\
        Gemma-2B & 86.41 & 86.16 & 86.38 \\
        Gemma-7B & 81.06 & 81.02 & 81.02 \\
        Meta-Llama-3-8B & 82.73 & 83.12 & 83.05 \\
        Falcon-2-11B & 86.18 & 86.12 & 86.12 \\ \midrule
        SmolLM-1.7B & 86.66 & 86.64 & 86.67 \\
        Gemma-2B-I & 87.96 & 88.01 & 87.91 \\
        Mistral-7B-I-v0.3 & 88.29 & 88.29 & 88.29 \\
        Gemma-7B-I & 85.14 & 85.15 & 85.15 \\
        Meta-Llama-3-8B-I & 84.11 & 84.12 & 84.18 \\ \bottomrule
        \end{tabular}
    \caption{Mean BERTScore recall values of outputs using greedy, top-k and top-p decoding with best instructions.}
    \label{tab:sampling_comparison}
\end{table*}
\section{Qualitative Analyses of Generated Outputs}

In this appendix section, we will do some qualitative analyses of the generated outputs by Language Models.

\subsection{Qualitative Examples for Mistral-7B-I}
\label{app:qual_examples}

We have a wide number of varying parameters. Showing outputs for 10 LMs (+3 SOTA), 8 prompt styles, 12 task types, 12 domains, 10 reasoning types is not practically feasible. However, since the dataset is public and we are using openly available LMs, we think any desired output is fairly reproducible. We still show some of the qualitative examples in Table~\ref{tab:examples} for reference for Mistral-7B-I-v0.3 on the prompt style with 8 examples and added task definition. We have only included the task instance, and removed the full prompt for brevity.

We see that in general, the outputs of the model are aligned and can be used directly. This is probably expected since it has a BERTScore recall value of 93.76, and Rouge-L value of 35.55 with the gold-standard label. For classification tasks also, it is generating the response that is perfectly aligned. We still have tried to find and outline some cases where the output is not perfect. This highlights that the model is instruction-tuned on a wide variety of dataset and is very powerful to use directly.

\begin{table*}[htbp]
    \centering
        \begin{tabular}{p{0.1\textwidth}|p{0.11\textwidth}|p{0.08\textwidth}|p{0.22\textwidth}|p{0.13\textwidth}|p{0.13\textwidth}}
        \toprule
        \textbf{Task Type} & \textbf{Domain} & \textbf{R. Type} & \textbf{Instance} & \textbf{Candidate} & \textbf{Reference} \\ \midrule
        Question Rewriting & Government and Politics & -- & What did this president major in during their college studies? How did the stock market perform while this president was in office? &  Did the stock market perform well while this president was in office, despite their college major being in economics? & Did this president major in economics or business? \\ \midrule
        Word Analogy & Common sense & Common sense, Analogical & cut : scissors. crack : ? & hammer & nutcracker \\ \midrule
        Grammar Error Correction & English Exams & -- & She know just that what teacher has told her but no more . &  She knows just what the teacher told her, but no more. & She know just what her teacher has told her but no more .  \\ \midrule
        Overlap Extraction & Natural Science & -- &  Sentence1: satellites are used to take images from space.
        Sentence2: satellites are used to take Meteorologists. &  satellites. & satellites \\ \midrule
        Cause-effect Classification & Web & Causal  & I received a package in the mail. (A) The package triggered my curiosity. (B) I took the package to the post office., Question: effect & A & A \\ \midrule
        Title Generation (choosing best title here) & Narrative, Story & Deduct-ive & Sentence 1: Crystal decided to learn how to sew. Sentence 2: She worked hard at cutting out the pattern. Sentence 3: Then she lined it up and sewed the edges up. Sentence 4: She had created a beautiful pillowcase! Sentence 5: Crystal decided to try a quilt next. Choices: a. The Sewing Machine. b. The Ram.
         & a & a \\ \bottomrule
    \end{tabular}
    \caption{Examples of generated responses (candidate) by \textbf{Mistral-7B-I} along with reference and task type, domain, reasoning type (R. Type). We have only included the instance (not the complete prompt) for brevity. What has to be done can be understood from the task type.}
    \label{tab:examples}
\end{table*}

\subsection{Explaining Incorrect Responses of Pre-trained models}
\label{app:pt_model_limitations}

In Section~\ref{sec:comparison_sota} and Appendix~\ref{app:aspects_lm_performance}, we observed that even the best pre-trained models are not able to match the performance of IT models on SOTA models. While there was a theoretical reasoning, in this appendix, we will take some qualitative observation of outputs and focus on explainability of errors for the two best performing pre-trained models - Gemma-2B and Falcon-2-11B using the best prompt style.

\begin{table*}[htbp]
    \centering
        \begin{tabular}{p{0.15\textwidth}|p{0.3\textwidth}|p{0.15\textwidth}|p{0.15\textwidth}|p{0.1\textwidth}}
        \toprule
        \textbf{Task Type /Domain/ Reasoning Type} & \textbf{Task Instance} & \textbf{Candidate} & \textbf{Reference} & \textbf{Identified Reason} \\ \midrule
        Coreference Resolution/ Commonsense/ Commonsense Reasoning & You must fill in the blank in a provided question with one of the two objects mentioned in the question. \textbf{...(truncated)} 
        
        Input: Bob finished eating his zuchinnis but couldn't eat all his watermelons, because there was a small amount of the \_ . & the answer is: & zuchinnis & not able to answer \\ \midrule
        Word Analogy/ Commonsense/ Relational & mainland : boat. portland : ? & <strong> car </strong> & car & output not aligned \\ \midrule
        Coreference Resolution/ Commonsense/ Commonsense Reasoning & Your job is to assess if a hypothesis sentence implies, contradicts, or is neutral with a given premise sentence. Respond with ""Contradiction"", ""Neutral"", or ""Entailment"" based on the relationship between the premise and hypothesis.
        
        Input: Premise: TORONTO, March 7 (Reuters) - The Canadian dollar weakened to a session low against the greenback after data showed the domestic economy unexpectedly shed jobs in February. \textbf{...(truncated)} <sep> Hypothesis: Toronto is the most populous city in Canada. & Neutral & Entailment & incorrect output \\ \midrule
        Answerability Classification/ Nutrition/ Reasoning on Social Interactions & Given a paragraph about cooking, and a set of conversational questions and answers about the paragraph, say whether the passage contains sufficient information to answer the follow-up question. \textbf{...(truncated)}

        Input: CONTEXT: Yes that's exactly what it means, apply salt to the steak. You shouldn't coat it, but you should apply salt very liberally. \textbf{...(truncated)} <sep> Q: What does it mean to 'salt' a steak prior to frying? & 100\% & YES & answering but differently \\
         \bottomrule
    \end{tabular}
    \caption{Examples of some responses (candidate) different from target by \textbf{Gemma-2B} using 4 examples with definition. We have only included the instance and some description (not the complete prompt), and also truncated the generated text for brevity. Full description can be viewed in the dataset.}
    \label{tab:error_analysis_gemma_2b}
\end{table*}

Table~\ref{tab:error_analysis_gemma_2b} outlines four different types of common errors observed in Gemma-2B with the best prompt style (4 examples with definition). We can see that in the second and fourth example, the model is able to answer the question. But, in the second example, it is adding extra HTML tags. In the fourth, instead of answering yes and no, it is generating `100\%'. Particularly, we found significant instances where outputs had extra HTML tags of \texttt{<strong>}, \texttt{<b>}, etc., despite the model getting 4 in-context examples to understand desired response. So, it can be inferred that Gemma-2B has a limitation of not being able to generate aligned responses learning from examples, and adding extra HTML tags to it. This is not observed for Gemma-2B-I; therefore, adapting the model for a specific application can eliminate such issues. In case 1, the model didn't generate any valid answer, and in case 3, it gave a wrong answer. In these scenarios, it is not certain if these issues can be resolved or are just limitations of the model's knowledge.

\begin{table*}[htbp]
    \centering
        \begin{tabular}{p{0.15\textwidth}|p{0.25\textwidth}|p{0.2\textwidth}|p{0.15\textwidth}|p{0.1\textwidth}}
        \toprule
        \textbf{Task Type /Domain/ Reasoning Type} & \textbf{Task Instance} & \textbf{Candidate} & \textbf{Reference} & \textbf{Identified Reason} \\ \midrule
        Natural Science/Overlap Extraction & Generate an overlapping word between the given two sentences. When you find the overlapping words, they don't have to match exactly, \textbf{...(truncated)}.

        Input: Sentence1: a warm front causes cloudy and rainy weather. 
        Sentence2: Cloudy, cool, or rainy weather can cause some algae scums to die. & 
        The overlapping word between the given two sentences is "weather". & cloudy & produces a different but correct output, but is not aligned \\ \midrule
        Answerability Classification/ News & You are given a sentence and a question in the input. If the information provided in the sentence is enough to answer the question, label ""Yes"", otherwise label ""No"". Do not use any facts other than those provided in the sentence while labeling ""Yes"" or ""No"". There are only two types of valid responses: Yes and No.
        
        Sentence: They didn't scream if they saw a dog, cat or chicken. 
        Question: What was Bobby's favorite thing? & To complete this task, I will analyze the given sentence and question to determine if the information provided in the sentence is enough to answer the question.
        
        Sentence: They didn't scream if they saw a dog, cat or chicken.
        Question: What was Bobby's favorite thing?
        Step 1: Identify the relevant information in the sentence.
        \textbf{...(truncated)}
        Output: No. & No & output not aligned \\ \midrule
        Coreference Resolution/ Commonsense/ Commonsense Reasoning & You need to answer a given question containing a blank (\_). Your answer must be one of the two objects mentioned in the question,  \textbf{...(truncated)}
        
        After the bike accident, the handle bar needed to be repaired, but the rim was fine, because the \_ misshapen. & The handle bar needed to be repaired, but the rim was fine, because the \_ misshapen.
        
        The blank is ""rim""." & handle bar & incorrect, not aligned \\
         \bottomrule
    \end{tabular}
    \caption{Examples of some responses (candidate) different from target by \textbf{Falcon-2-11B} using 8 examples with definition. We have only included the instance and some description (not the complete prompt), and also truncated the generated text for brevity. Full description can be viewed in the dataset.}
    \label{tab:error_analysis_falcon_2_11b}
\end{table*}

The generated outputs for Falcon-2-11B, as given in Table~\ref{tab:error_analysis_falcon_2_11b} was found to have other kinds of differences. First, no HTML tags were witnessed, which also confirms that it was specific to Gemma-2B. In Falcon-2, the outputs were often given as sentences, like Example 1 and Example 3 from the table. Example 1 has a correct answer, but it does not match the reference. However, while the output is misaligned, it is not wrong. For Example 3, the output is both misaligned and incorrect. There were several outputs that were like this. But, there were even more cases like the second example, where the model generated a sequence of steps for itself before giving the result, something like COT prompting~\citep{cot_prompting}. The result was correct ultimately. This case can be easily handled by aligning the output, or post-processing it to extract desired text. We observed that ignoring these differences, the outputs of Falcon-2-11B were generally correct, making it a very powerful model if used appropriately. We couldn’t compare it to the IT version, as it is not available yet. 
\section{Implementation Details}
\label{app:implementation_details}

We used a publicly available dataset Super Natural Instructions~\citep{super_natural_instructions} for this work. It dataset is a meta-dataset created using multiple datasets. The paper reports its creation steps and multi-stage quality control process including automatic and manual processes, which were sufficient to eliminate the risks of personal or offensive content. We thoroughly went through the dataset paper, its collection process, and manually examined few samples of the dataset to verify this.

We use a single Nvidia A-40 GPU with 48 GB GPU memory to conduct all our experiments on a GPU cluster for each run. We define one run as a single forward pass on one model using a single prompt style. The batch sizes used are different and range from 2-8 for different models based on their sizes (2 for 11B model, 4 for 7B models, 8 for 2B and 3B models). Each run varied from approximately 80 minutes (for Gemma-2B-I) to approximately 60 hours (for Falcon-2-11B). 

All model implementations are used from HuggingFace~\footnote{\href{https://huggingface.co}{https://huggingface.co}}, except the SOTA models for which we use the OpenAI APIs~\footnote{\href{https://openai.com/index/openai-api/}{https://openai.com/index/openai-api/}}. We have used all artifacts as per their intended use. The implementations used and license details are provided in Table~\ref{tab:artifacts}.

All inputs were tokenized using respective model tokenizers and left-padded to match the context size of each model, and 512 max new tokens were generated during inference by the model.

We perform all inferences with 4-bit quantized~\citep{dettmers2023qlora} versions of all models using Huggingface BitsAndBytes, along with Flash Attention 2~\citep{dao2022flashattention}.

As discussed before, we are also sharing a GitHub repository of our implementation (link available on page 1 footnote) as a utility which will allow evaluating any LM using this dataset and generating these visualizations. It also supports doing this using other evaluation metrics discussed in Table~\ref{tab:all_metrics_results} if required.

\begin{table*}[htbp]
    \centering
    \begin{tabular}{p{0.33\textwidth}|p{0.3\textwidth}|p{0.33\textwidth}}
        \toprule
        \textbf{Artifact} & \textbf{Implementation Link} & \textbf{License}  \\
        \midrule
        Super Natural Instructions Dataset & \href{https://instructions.apps.allenai.org}{Page (v2.8 used)} & \href{https://github.com/allenai/natural-instructions/blob/master/LICENSE}{Apache 2.0 License} \\ \midrule
        SmolLM-1.7B & \href{https://huggingface.co/HuggingFaceTB/SmolLM-1.7B}{Model Card} & \href{https://www.apache.org/licenses/LICENSE-2.0}{Apache 2.0 License} \\
        Gemma-2B & \href{https://huggingface.co/google/gemma-2b}{Model Card} & \href{https://ai.google.dev/gemma/terms}{Apache 2.0 License} \\
        Gemma-2-2B & \href{https://huggingface.co/google/gemma-2-2b}{Model Card} & \href{https://ai.google.dev/gemma/terms}{Apache 2.0 License} \\
        Mistral-7B-v0.3& \href{https://huggingface.co/mistralai/Mistral-7B-v0.3}{Model Card} & \href{https://huggingface.co/datasets/choosealicense/licenses/blob/main/markdown/apache-2.0.md}{Apache 2.0 License} \\
        Gemma-7B & \href{https://huggingface.co/google/gemma-7b}{Model Card} & \href{https://ai.google.dev/gemma/terms}{Apache 2.0 License} \\
        Meta-Llama-3-8B & \href{https://huggingface.co/meta-llama/Meta-Llama-3-8B}{Model Card} & \href{https://github.com/meta-llama/llama3/blob/main/LICENSE}{Meta Llama-3 Community License} \\
        Falcon-2-11B & \href{https://huggingface.co/tiiuae/falcon-11B}{Model Card} & \href{https://falconllm.tii.ae/falcon-2-terms-and-conditions.html}{Falcon 2 11B TII License} \\
        SmolLM-1.7B-I & \href{https://huggingface.co/HuggingFaceTB/SmolLM-1.7B-Instruct}{Model Card} & \href{https://www.apache.org/licenses/LICENSE-2.0}{Apache 2.0 License} \\
        Gemma-2B-I & \href{https://huggingface.co/google/gemma-2b-it}{Model Card} & \href{https://ai.google.dev/gemma/terms}{Apache 2.0 License} \\
        Gemma-2-2B-I & \href{https://huggingface.co/google/gemma-2-2b-it}{Model Card} & \href{https://ai.google.dev/gemma/terms}{Apache 2.0 License} \\
        Mistral-7B-I-v0.3 & \href{https://huggingface.co/mistralai/Mistral-7B-Instruct-v0.3}{Model Card} & \href{https://huggingface.co/datasets/choosealicense/licenses/blob/main/markdown/apache-2.0.md}{Apache 2.0 License} \\
        Gemma-7B-I & \href{https://huggingface.co/google/gemma-7b-it}{Model Card} & \href{https://ai.google.dev/gemma/terms}{Apache 2.0 License} \\
        Meta-Llama-3-8B-I & \href{https://huggingface.co/meta-llama/Meta-Llama-3-8B-Instruct}{Model Card} & \href{https://github.com/meta-llama/llama3/blob/main/LICENSE}{Meta Llama-3 Community License} \\ \midrule
        METEOR & \href{https://www.nltk.org/_modules/nltk/translate/meteor_score.html}{Doc} & \href{https://huggingface.co/spaces/evaluate-metric/meteor/blame/d33847fd9d688beb98d7577c2960b006d361336a/meteor.py}{Apache 2.0 License} \\
        ROUGE & \href{https://github.com/google-research/google-research/tree/master/rouge}{Doc} & \href{https://github.com/google-research/google-research/blob/master/LICENSE}{Apache 2.0 License} \\
        BERTScore & \href{https://github.com/Tiiiger/bert_score}{Doc (used using Roberta Large)} & \href{https://github.com/Tiiiger/bert_score/blob/master/LICENSE}{MIT License} \\
        \bottomrule
    \end{tabular}
    \caption{Details of artifacts used with implementation links and license details.}
    \label{tab:artifacts}
\end{table*}

\end{document}